\documentclass{article} 
\usepackage[preprint]{colm2026_conference}

\usepackage{microtype}
\usepackage{hyperref}
\usepackage{url}
\usepackage{booktabs}
\usepackage{amsmath}

\usepackage{graphicx}
\usepackage{subcaption}  

\usepackage{pgfplots}
\pgfplotsset{compat=1.18}
\usepackage{pgfplotstable}
\usepackage{booktabs}
\usepgfplotslibrary{groupplots}
\usetikzlibrary{positioning}

\usepackage{lineno}

\definecolor{darkblue}{rgb}{0, 0, 0.5}
\hypersetup{colorlinks=true, citecolor=darkblue, linkcolor=darkblue, urlcolor=darkblue}

\title{Quantifying Trust: \\Financial Risk Management for Trustworthy AI Agents}

\author{%
\textbf{Wenyue Hua}\textsuperscript{1}\thanks{Contacts: Wenyue Hua: \texttt{wenyuehua@microsoft.com}, Chandler Fang:\texttt{cfang@t54.ai}. The views expressed in this paper are solely those of the authors and do not represent the views or opinions of any institution. Work initiated during Wenyue Hua's postdoctoral appointment at University of California, Santa Barbara.
}\quad 
\textbf{Tianyi Peng}\textsuperscript{2} \quad 
\textbf{Chi Wang}\textsuperscript{3} \quad 
\textbf{Jiaxin Pei}\textsuperscript{4} \quad \\
\textbf{Ian Kaufman}\textsuperscript{5} \quad
\textbf{Bryan Lim}\textsuperscript{6} \quad 
\textbf{Chandler Fang}\textsuperscript{\textbf{5}} \\
\textsuperscript{1}Microsoft Research \quad 
\textsuperscript{2}Columbia University \quad
\textsuperscript{3}Google DeepMind \quad
\textsuperscript{4}Stanford University \quad \\
\textsuperscript{5}t54.ai \quad
\textsuperscript{6}Virtuals ACP \quad \\
}

\usepackage[normalem]{ulem}

\begin{document}

\ifcolmsubmission
\linenumbers
\fi

\maketitle

\begin{abstract}

Prior work on trustworthy AI emphasizes model-internal properties such as bias mitigation, adversarial robustness, and interpretability. As AI systems evolve into autonomous agents deployed in open environments and increasingly connected to payments or assets, the operational meaning of trust shifts to end-to-end outcomes: whether an agent completes tasks, follows user intent, and avoids failures that cause material or psychological harm. These risks are fundamentally product-level and cannot be eliminated by technical safeguards alone because agent behavior is inherently stochastic. To address this gap between model-level reliability and user-facing assurance, we propose a complementary framework based on risk management. Drawing inspiration from financial underwriting, we introduce the \textbf{Agentic Risk Standard (ARS)}, a payment settlement standard for AI-mediated transactions. ARS integrates risk assessment, underwriting, and compensation into a single transaction framework that protects users when interacting with agents. Under ARS, users receive predefined and contractually enforceable compensation in cases of execution failure, misalignment, or unintended outcomes. This shifts trust from an implicit expectation about model behavior to an explicit, measurable, and enforceable product guarantee. We also present a simulation study analyzing the social benefits of applying ARS to agentic transactions. ARS's implementation can be found at \url{https://github.com/t54-labs/AgenticRiskStandard}.

\end{abstract}

\section{Introduction}

Trustworthy AI~\citep{li2023trustworthy} has become a central theme in modern artificial intelligence, with many research directions addressing different aspects of safety and reliability. Work on bias detection and mitigation~\citep{lin2025investigating, kumar2024decoding} seeks to reduce harmful stereotypes and disparate impacts in model outputs. Robustness and red teaming~\citep{chen2024agentpoison, ge2024mart} evaluate system behavior under adversarial attacks~\citep{kumar2023certifying}, distribution shifts~\citep{ren2025llms}, and prompt-based manipulation~\citep{liu2023prompt, greshake2023not}. Explainability~\citep{bilal2025llms, mumuni2025explainable} and mechanistic interpretability~\citep{singh2024rethinking, dunefsky2024transcoders} attempt to characterize the internal mechanisms of models, opening the ``black box''~\citep{gat2023faithful, bhattacharjee2023towards} to better understand how outputs are produced. Together these research directions have improved the technical reliability and transparency of AI systems~\citep{ahn2024impact, cau2023effects}.

As AI systems move from laboratory prototypes~\citep{devlin2019bert, liu2019roberta} to widely deployed infrastructure~\citep{brown2020language, comanici2025gemini, yang2025qwen3, liu2024deepseek, kwon2023efficient, zheng2024sglang}, the practical meaning of trust increasingly depends on end-to-end outcomes during real-world deployment. Large language models enable agentic systems~\citep{masterman2024landscape} that write and execute code, invoke external tools~\citep{patil2025berkeley}, perform multi-step reasoning~\citep{kim2024husky, aksitov2023rest}, maintain memory~\citep{hu2025memory, kang2025memory}, and interact with external environments~\citep{zhao2024epo}. These systems are increasingly offered as commercial services with subscription pricing, usage-based pricing, or per-task fees. In this setting, agents begin to function as economic actors that automate workflows~\citep{tang2025autoagent, schwartz2023enhancing}, provide delegated digital labor~\citep{chen2026trajectory, de2026openclaw, vella2024novice}, and execute financial operations such as trading~\citep{yang2025survey}. 

\begin{figure}
    \centering
    \includegraphics[width=0.85\linewidth]{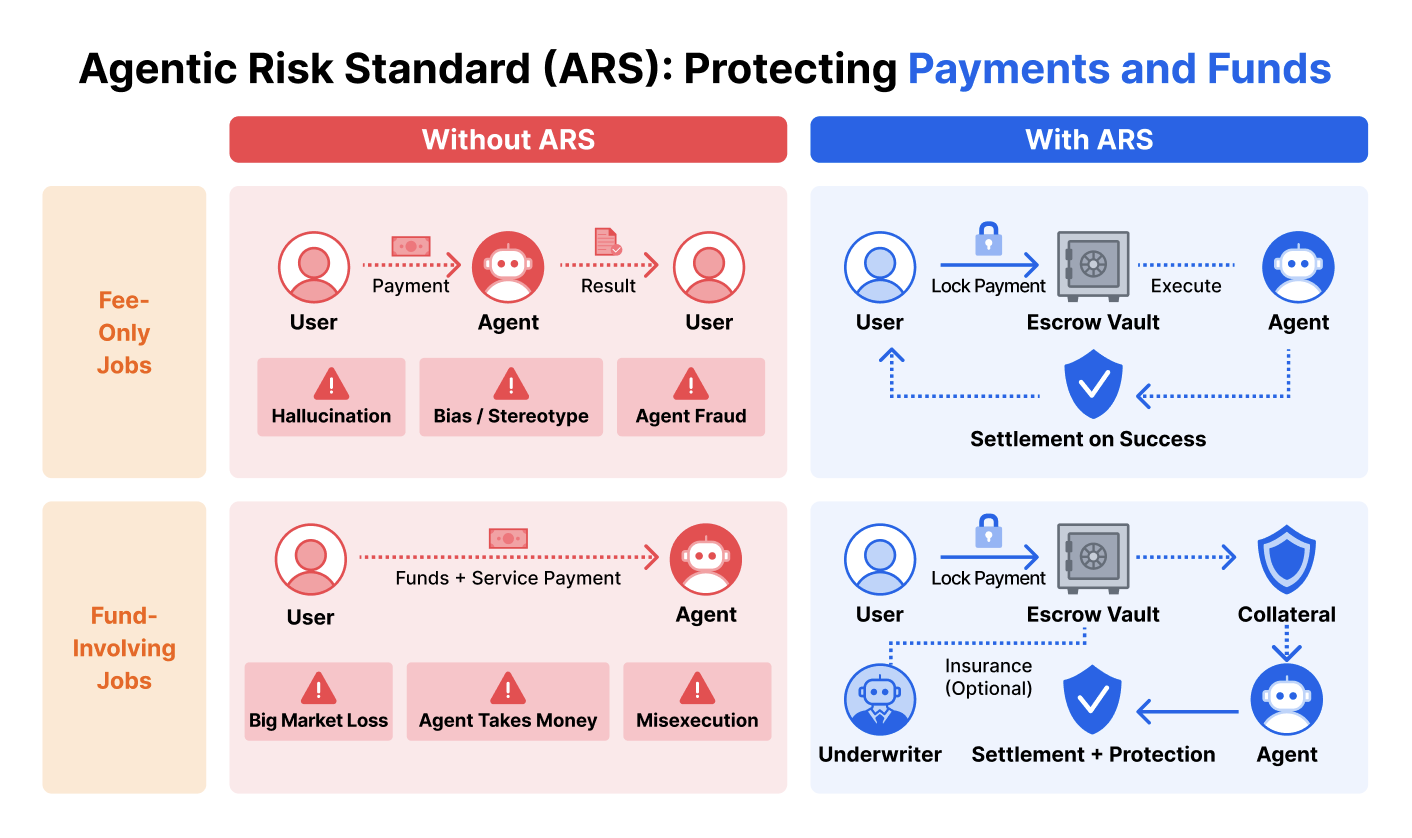}
    \caption{\textsc{ARS} is a transaction-layer assurance standard for agentic services that converts stochastic, outcome-level risk into explicit settlement rules. Without \textsc{ARS}, users must prepay agents (and in fund-moving tasks, also hand over execution capital), exposing them to non-delivery, misexecution, and downstream harms. With \textsc{ARS}, service fees are locked in an escrow vault and released only upon successful evaluation; for fund-moving tasks, users can optionally purchase underwriting coverage while the service provider posts collateral before principal is released. This shifts trust from model-internal reliability to auditable, enforceable guarantees over product-level outcomes.}
    \label{fig:main}
\end{figure}

As a result, the risks faced by users increasingly stem from product-level failures such as non-delivery, misexecution, misalignment with user intent, financial loss, and other downstream harms that are not adequately captured by model-internal evaluation metrics.

This deployment shift exposes a fundamental guarantee gap. Although alignment and robustness methods can reduce the likelihood of harmful behavior, they do not eliminate failure risk entirely~\citep{yang2023shadow, fang2026safethinker}. This limitation is reinforced by a growing body of work showing that existing safety mechanisms degrade under adversarial prompting and other stressors~\citep{zou2023universal, wang2025adversarial}. More fundamentally, large language models and vision-language models are inherently stochastic~\citep{evstafev2025paradox}, so no training procedure can fully remove the possibility of failure. As a result, technical safeguards can offer only probabilistic reliability, whereas users in high-stakes settings often require enforceable guarantees over outcomes. This tension becomes especially visible in real-world applications:

\begin{enumerate}
\item \textbf{Coding agent.} A developer uses an AI coding assistant to modify a production codebase. Even a subtle error can propagate into service outages, data corruption, or costly recovery work, creating consequences far beyond a localized model mistake.

\item \textbf{Tax-filing agent.} An agent prepares and submits a tax filing on behalf of a user. A filing error can trigger penalties, audits, delayed reimbursements, or legal disputes, turning a single model failure into a broader financial and administrative burden.

\item \textbf{Customer-service agent.} A company deploys an AI agent to handle customer inquiries. An incorrect response or unauthorized commitment can escalate into contractual disputes, regulatory exposure, or legal liability, with consequences that extend well beyond the original interaction.

\item \textbf{Financial transaction agent.} A user delegates a currency exchange or trading action to an autonomous agent~\citep{xiao2024tradingagents}. Because the agent acts directly on real assets, a single failure can produce immediate realized loss rather than a contained degradation in model performance.
\end{enumerate}

In each case, delegation creates downside risk that may be significantly larger than the direct price of the service or the expected benefit from automation. The core tension is that users must grant the agent pre-verification access to valuable assets such as funds, codebases, credentials, datasets, or operational permissions. Once such access is granted, failure may result not only in non-delivery, but also in corruption, disclosure, misuse, or irreversible commitment of pre-existing assets, making total loss difficult to bound ex ante. When potential loss cannot be bounded ex ante, rational users are likely to restrict delegation to low-stakes tasks. This asymmetry between limited upside and potentially large downside creates adoption friction and slows the deployment of agent-based services\footnote{We focus on monetary assets for a clarify of loss model, while extension to non-monetary assets is straightforward but is highly context-dependent and personalized.}.

\textit{Traditional commerce has long relied on risk-allocation mechanisms to support transactions with uncertain outcomes.} Across domains such as construction, professional services, financial markets, e-commerce, and decentralized finance, mechanisms such as escrow, collateralization, insurance, and clearinghouses are used to bound downside risk and enable participation under uncertainty. These mechanisms do not eliminate operational failure. Instead, they define how risk is allocated, when funds are released, and how losses are compensated when things go wrong. Despite the growing deployment of AI agents, there is still no standardized settlement standard for agentic transactions. In most current deployments, the user simply pays the service fee upfront and then bears the residual execution risk of the agent. This is a different axis of safety from model alignment or robustness. Model-level safeguards aim to reduce the probability of failure; settlement-layer safeguards determine how trust, payment, and compensation are handled when failures remain possible. 

\begin{table*}[t]
\centering
\small
\begin{tabular}{p{2.5cm} p{2cm} p{3cm} p{3.3cm} p{2.3cm}}
\toprule
\textbf{Industry / Context} & \textbf{Service Provider} & \textbf{Primary Risk} & \textbf{Risk Protection Mechanism} & \textbf{Who Bears Residual Risk} \\
\midrule
Construction & Contractor & Project failure, delay, quality defects & Performance bond & Insurer / surety provider \\

Professional services & Doctor, lawyer, accountant & Professional error, malpractice & Liability insurance & Insurance provider \\

E-commerce platforms & Online seller & Non-delivery, poor service & Escrow payment & Platform escrow account \\

Government procurement & Contractor & Contract non-performance & Bid bond / performance bond & Insurance / surety \\

Financial markets & Traders / brokers & Default and market volatility & Margin requirements + clearinghouse & Clearinghouse \\

DeFi / crypto lending & Borrower & Loan default & Collateralization & Smart contract protocol \\

Sharing economy & Drivers, hosts & Service disputes, property damage & Deposit + platform insurance & Platform / provider \\

\textbf{AI agents (ARS)} & AI Agent & Misexecution, hallucination, jailbreaking & Escrow + collateral + underwriting & Underwriter + collateral pool \\

\bottomrule
\end{tabular}
\caption{\textbf{Risk protection mechanisms across industries.} 
High-stakes services rarely rely on trust alone. Instead, industries introduce financial mechanisms that allocate risk across participants. \textsc{ARS} applies the same principle to AI agents by combining escrow settlement, collateralization, and optional underwriting to provide outcome-level guarantees.}
\label{tab:risk-mechanisms}
\end{table*}

To address this gap, we introduce the \textbf{Agentic Risk Standard (\textsc{ARS})}, a settlement-layer standard for agentic services. \textsc{ARS} specifies how payments, collateral, claims, and reimbursement are handled over the lifecycle of an agentic job. Rather than prescribing model behavior, \textsc{ARS} defines settlement semantics through a structured state machine governing actions such as escrow locking and release, collateral posting, claim filing, dispute resolution, and reimbursement. All financially relevant transitions depend on auditable signals that verify task outcomes. In this way, stochastic task execution can be mapped into deterministic settlement rules.

\textbf{A central design goal of \textsc{ARS} is modularity.} The standard does not assume a universal model for loss quantification, premium pricing, or underwriting decisions, since these depend on the application domain, the delegated asset, and the task-specific structure of harm. Instead, \textsc{ARS} provides a standardized transaction and settlement interface into which domain-specific risk models can be integrated. This makes \textsc{ARS}'s semantic applicable across heterogeneous agentic tasks while separating settlement logic from risk estimation logic. In addition, we evaluate \textsc{ARS} in a simulated environment in which users, service providers, and underwriters interact through the protocol. Rather than proposing a universal pricing model, the simulation studies how the settlement design behaves under varying externally supplied risk and contract parameters, and analyzes the resulting tradeoffs among user protection, provider participation, and market adoption. The standard specification is released as open source to support adoption and further research on transaction-layer guarantees for AI agents.

In summary, this paper makes two contributions. First, it identifies a transaction-layer dimension of agent safety that is orthogonal to model-internal safety and robustness, centered on enforceable guarantees over end-to-end outcomes. Second, it introduces \textsc{ARS}, a payment settlement standard for agentic services that formalizes how escrow, collateral, claims, and reimbursement should be handled when users delegate tasks to AI agents.

\section{Related Work}

\paragraph{Agent architectures, tool use, and environment interaction.}
A growing body of work studies how to extend foundation models into agentic systems that can plan and act in external environments. Tool-augmented agents integrate APIs and symbolic tools to execute structured actions beyond free-form generation~\citep{qin2023toolllm}. Complementary efforts standardize how tools and context are exposed to models through open interfaces, including Model Context Protocol (MCP)-style designs~\citep{guo2025mcp}, and organize reusable capabilities through skill-oriented architectures. In parallel, GUI-based agents operationalize as interface operators that perceive screens and execute actions across web and application environments~\citep{deng2023mind2web, nong2024mobileflow, zhang2024large, zhang2025appagent, sager2025comprehensive}. These lines of work primarily address capability, interoperability, and execution in complex environments, but they do not by themselves specify how economic liability is allocated when execution is incorrect or incomplete.

\paragraph{Trustworthy AI and agent safety.}
Trustworthy AI research addresses fairness, robustness, interpretability, and alignment, with extensive work on bias detection and mitigation, adversarial robustness and red-teaming, mechanistic interpretability, and alignment training~\citep{borah2024towards, chen2024agentpoison, bereska2024mechanistic, ji2023beavertails}. While these approaches improve technical reliability and reduce harmful behavior, they generally provide probabilistic guarantees at the model level. For agentic systems, especially those interacting with external tools and markets, residual uncertainty remains at the level of task outcomes and economic consequences, motivating additional mechanisms that operate at the product or transaction layer.

\paragraph{Agent coordination and commerce protocols.}
Several emerging protocols aim to standardize interoperability and commerce for agents. MCP specifies a standardized interface for connecting model hosts to external tools and data sources, emphasizing a common protocol surface for tool invocation and context exchange. \emph{Universal Commerce Protocol (UCP)}~\citep{google_ucp_2026} proposes a common language and primitives for agentic commerce spanning consumer surfaces, merchants, and payment providers, with the goal of interoperable shopping and transaction flows. In decentralized settings, \emph{Agent Commerce Protocol (ACP)}~\citep{virtuals_acp_2026} defines a smart-contract–mediated workflow for agent transactions that emphasizes verifiable agreements and an explicit evaluation phase to assess outcomes against agreed terms. These efforts are most directly concerned with interoperability, transaction coordination, and settlement plumbing; they do not focus on actuarial risk pricing or underwriting mechanisms that explicitly allocate financial liability under stochastic execution.

\paragraph{Finance-inspired risk allocation and assurance.}
Finance and insurance provide mature mechanisms for managing uncertainty through risk measurement, pricing, and underwriting~\citep{mcneil2015quantitative, embrechts2000actuarial, meyers1999underwriting, briys2001insurance}. Related instruments such as performance bonds and escrow-based settlement, tie payment to verifiable outcomes and contractually allocate downside risk~\citep{compton2004performance, bessembinder2008measuring, artemenkov2025financial}. Recent discussions on AI governance and liability predominantly emphasize legal and regulatory structures for accountability~\citep{hacker2024sustainable, hacker2023ai, anderljung2023frontier}, leaving open the system question of how to embed quantitative underwriting and enforceable economic guaranties directly into agent execution protocols.

Our work connects these threads by introducing \textsc{ARS} as a settlement-layer service for agentic services. \textsc{ARS} augments existing coordination and commerce standards with explicit fund-control semantics: conditional escrow release, conditional collateralization, underwriting approval, and structured claims and dispute handling. \textbf{\textsc{ARS} transforms stochastic task execution to deterministic settlement outcomes.} In this way, \textsc{ARS} complements model-level trustworthy AI by providing transaction-layer assurance with explicit liability allocation and enforceable user protection.

\section{Preliminaries: Financial Risk Management Concepts}

ARS draws on a set of established concepts from financial risk management and insurance. This section provides concise definitions of the core terms used throughout the paper, grounding the standard's design in their financial-domain meanings.

\paragraph{Escrow} is a conditional custody arrangement where a third party holds funds or assets on behalf of transacting parties and releases them only when predefined conditions are satisfied. In a standard commercial escrow, a buyer deposits payment into the escrow account before receiving a good or service; and the escrow is only released to the seller upon verified delivery, and refunds the buyer otherwise. Escrow eliminates direct counterparty exposure by ensuring that neither party can unilaterally access the held funds: the seller cannot receive payment before delivery, and the buyer cannot reclaim funds without a valid refusal trigger. 

\paragraph{Underwriting} is the process by which a risk-bearing party such as the \emph{underwriter} evaluates, prices, and assumes a specified risk in exchange for compensation. In financial and insurance markets, an underwriter assesses the probability and magnitude of a loss event, sets terms under which it will absorb that loss, and charges premium for doing so. By accepting risk contractually, the underwriter converts uncertain outcomes into determinate liabilities with explicit remedies.

\paragraph{Premium} is the price paid by the insured party to the underwriter in exchange for coverage against a specified loss event. It is typically set to reflect the underwriter's expected loss exposure which is the \emph{actuarially fair premium} plus a loading factor that accounts for risk capital, administrative costs, and profit margin. 

\paragraph{Collateral} is an asset pledged by one party to a transaction as security against the risk of non-performance or loss. In lending and derivatives markets, collateral reduces counterparty exposure by ensuring that a portion of potential losses can be recovered directly from the pledged assets without recourse to legal proceedings. Usually, the collateral is locked in a custody vault and is either returned upon successful execution or forfeited when a covered failure occurs, offsetting the underwriter's reimbursement obligation.

\section{Assurance Modes by Task Type}
\textsc{ARS} models each agentic task as a \emph{job} with a signed agreement, a controlled lifecycle, and post-execution evaluation. The key design principle is that the assurance mechanism should match the job's \emph{pre-verification fund (asset) exposure}: whether the job requires releasing users' pre-existent fund or granting authority to the provider agent to perform a task before the outcome can be verified. Recall the examples from Section~1: in tax-filing (Example~2), the user's exposure of loss is limited to the service fee that's predestined to pay to the agent service; in the financial-transaction scenario (Example~4), the agent must have control access user capital before delivery. We call these \textbf{fee-only tasks} and \textbf{fund-involving tasks}, respectively.

\paragraph{Fee-only tasks} have no pre-verification fund exposure (e.g., generating slide decks~\citep{liang2025slidegen, liangpaper2slide}, images~\citep{zhang2025postergen, shin2025postermate}, music~\citep{yu2023musicagent, deng2024composerx}, and reports~\citep{tian2025template}). The dominant risk is non-delivery or defective delivery, observable after execution. \textsc{ARS} adopts \textbf{escrow-based conditional settlement}: payment is locked and released only upon verified delivery, aligning with escrow patterns in prior agent commerce workflows such as Virtuals ACP.

\paragraph{Fund-involving tasks} require releasing a \emph{principal} or granting financial authority \emph{before} outcomes can be verified (\emph{e.g.}, trading~\citep{xiao2024tradingagents, ding2024large, yu2025finmem, li2024cryptotrade, qian2025agents}, leveraged positions, financial API calls). Escrow alone is insufficient because the dominant exposure occurs prior to evaluation. \textsc{ARS} adds \textbf{underwriting-based assurance}: a risk-bearing party prices the outcome risk, may require provider collateral, and commits to reimbursement under explicit failure triggers, while the service fee is still escrowed separately.

Both modes share the same job abstraction and phase transitions but differ in settlement semantics: escrow enforces \emph{conditional release after verification}; underwriting enforces \emph{risk-gated release with contingent reimbursement}.

\section{Agentic Risk Standard}

This section specifies \textsc{ARS} as a \emph{job-level state machine} with explicit actions and authorization conditions. The goal is to make execution unambiguous: at any point, a job is in a well-defined state; only a small set of typed actions are permitted from that state; and financially relevant transitions such as locking funds, releasing funds, and paying claims occur only when their preconditions are satisfied.

\subsection{Roles, Job, and Structured Agreement}

\textsc{ARS} involves three primary roles: a \textbf{requestor} which is a human user or user-side assistant that creates a job and provides payment intent, a \textbf{business agent} which is the service provider that executes the job, and an \textbf{underwriter} that prices and assumes specified outcome risk for jobs that require pre-verification fund exposure. Depending on deployment, \textsc{ARS} may also include an \textbf{evaluator/arbiter} responsible for delivery verification and dispute adjudication, and an optional \textbf{override signer} which has to be the human user who can authorize exceptional transitions under policy-defined conditions.

A \emph{job} is the unit of execution and settlement, uniquely identified by \texttt{job\_id}. Each job is anchored by a \textbf{structured agreement}, a signed object that specifies: (1) the task statement such as action or deliverable with parameters and constraints, (2) the assurance mode which is \emph{fee-only} or \emph{fund-involving}, (3) payment amount and custody rules, (4) acceptance criteria and evaluation procedure, (5) deadlines and time windows such as delivery, claim, and dispute. The structured agreement binds the parties and fixes the semantics of success and failure for the job.

\textsc{ARS} treats the structured agreement as the canonical reference for authorization and settlement: state transitions that lock or release funds, post collateral, or pay out claims are valid only when consistent with the agreement and supported by the required auditable signals.

\subsection{Authorization and Custody Components}
\textsc{ARS} distinguishes \emph{authorization} about who can approve a financially relevant transition from \emph{custody} about which component actually holds and moves funds. This separation makes settlement semantics explicit: signers emit auditable approvals and evidence, while custody components enforce conditional fund movement.

\paragraph{Two fund tracks: fee settlement vs.\ execution principal.}
\textsc{ARS} separates \emph{service compensation} from \emph{execution principal}. The \textbf{fee track} concerns the service fee paid to the business agent and is handled via escrow-style conditional settlement. The \textbf{principal track} concerns user funds that must be released pre-execution for fund-involving actions; this track introduces pre-verification exposure. Fee-only jobs instantiate only the fee track. Fund-involving jobs may instantiate both tracks: the fee track remains escrowed, while the principal track is optionally protected by underwriting if the requestor adopts coverage. When coverage is not adopted, \textsc{ARS} may still record principal release and execution evidence, but provides no reimbursement guarantee for principal loss.

\paragraph{Signers and auditable signals.}
\textsc{ARS} uses authenticated messages and signatures to authorize state transitions and to produce auditable signals for settlement. The main signers are the \textbf{requestor-side signers}, the \textbf{business agent}, the \textbf{underwriter}, and the \textbf{evaluator/arbiter}.

For the \textbf{fee track}, the requestor authorizes \texttt{LockFeeEscrow} to place the service fee into escrow. The business agent emits a signed delivery signal such as \texttt{SubmitDeliverable} for artifacts or \texttt{SubmitExecutionEvidence} for fund-involving actions that anchors an auditable reference for evaluation. The evaluator/arbiter emits \texttt{EvaluateOutcome}, which determines whether the escrowed fee is released or refunded.

For the \textbf{principal track} in fund-involving jobs, underwriting protection is \emph{optional} and is enabled only if the requestor accepts the quoted premium and the business agent agrees on the collateral. Both parties must bind an underwriting policy in the structured agreement. When coverage is adopted, pre-execution principal release is gated by underwriting decisions, optional collateral, and an adaptive authorization predicate that depends on whether the requestor is a human or a user-side assistant acting for a human. Let the requestor-side signer set be either $\{\textsf{H}\}$ or $\{\textsf{A},\textsf{H}\}$ where $\textsf{H}$ refers to human user and $\textsf{A}$ refers to agent. When the requestor is a human user, principal release requires approval from either the human requestor or the underwriter (a 1-of-2 rule). When the requestor is an assistant agent acting as proxy of human user, principal release requires the assistant's approval and one additional approval from either the human principal or the underwriter (a 2-of-3 rule with the assistant as a mandatory signer). The underwriter authorizes coverage binding with approve/reject, premium quote, and collateral requirements; the business agent authorizes collateral posting and submits execution evidence; and the evaluator/arbiter emits adjudicated outcome signals used for claim eligibility and collateral settlement.

\paragraph{Custody and fund-control semantics.}
\textsc{ARS} assumes a settlement layer that implements conditional custody and settlement actions: Fee escrow vault holds the service fee and enforces conditional \emph{release} or \emph{refund} based on auditable outcome signals. Collateral vault holds business agent collateral when required and enforces \emph{unlock} or \emph{slash} according to the collateral policy and adjudicated outcomes. Reimbursement payout vault is optional, which holds underwriter capital/reserves for enforceable claim payment. When used, eligible claims can be paid automatically under the bound policy; otherwise, payouts are executed from the underwriter treasury while still producing an auditable \texttt{payout\_ref}.

\paragraph{Payment rails and transfer receipts.}
Actual transfers are executed via underlying \textbf{payment rails}, such as on-chain transfers, exchange APIs, or traditional payment networks. Vault actions that move funds produce verifiable receipts, referenced in the standard as \texttt{*\_ref} (\emph{e.g.}, \texttt{lock\_ref}, \texttt{transfer\_ref}, \texttt{settlement\_ref}, \texttt{payout\_ref}), which serve as auditable evidence that a specific movement or custody change occurred.

\paragraph{Premium payment and opting out.}
Underwriting coverage must be explicitly adopted in the structured agreement. If underwriting terms are not accepted, \emph{i.e.}, the requestor declines the quoted premium, or the business agent refuses required collateral, the job transitions to a \emph{human-override} path in which the human requestor explicitly acknowledges and accepts full pre-execution risk. Upon a valid override signature, the standard may proceed with principal release under the non-underwritten semantics; otherwise, the job terminates.

\subsection{Lifecycle and State Model}
\textsc{ARS} specifies a job as a deterministic lifecycle governed by a structured agreement $\mathcal{A}$. At any time, a job is in a well-defined phase and a well-defined transaction state, where a phase refers to a coarse lifecycle stage shared by all jobs, and the transaction state is the fine-grained state that controls fund custody and authorization gates. For each state, only a restricted set of actions is allowed, and all fund movements are conditional on authenticated actions and auditable signals bound to \texttt{agreement\_hash}.

All jobs traverse the same high-level phases:
\[
\texttt{REQUEST} \rightarrow \texttt{NEGOTIATION} \rightarrow \texttt{TRANSACTION} \rightarrow \texttt{EVALUATION} \rightarrow \texttt{CLOSED}
\]
with \texttt{CANCELLED} as a terminal outcome reachable when any continuation conditions are not met. \texttt{REQUEST} creates a job intent, \texttt{NEGOTIATION} converges on the agreement $\mathcal{A}$ from both requestor side and the service side, \texttt{TRANSACTION} governs fund custody and authorized execution, \texttt{EVALUATION} determines outcome and settlement, and \texttt{CLOSED} finalizes escrow/collateral/payout actions.

During \texttt{TRANSACTION} phase, \textsc{ARS} separates a \emph{fee track} that escrows the service fee and settles it conditionally after evaluation, and  a \emph{principal track} that governs pre-execution release of execution principal. The fee track is always escrow-based. The principal track is either underwriting-protected when coverage is adopted via premium acceptance and policy binding or proceeds under a human-override acknowledgement in which the human requestor explicitly bears pre-execution risk.

ARS determinism is enforced by (1) agreement binding via \texttt{agreement\_hash}, (2) explicit authorization predicates for principal release, and (3) custody components that execute releases, refunds, collateral settlement, and payouts only when the corresponding preconditions are satisfied.

\subsection{ARS Actions}
The tables below define ARS's action vocabulary as typed, authenticated messages that drive the job state machine and fund-control rules across phases. Actions may be implemented as signed API calls, RPC messages, on-chain events, or smart-contract invocations; the specification only requires that each action is attributable to a standard role and bound to the signed structured agreement via \texttt{agreement\_hash}.

\begin{figure}[!ht]
    \centering
    \includegraphics[width=0.85\linewidth]{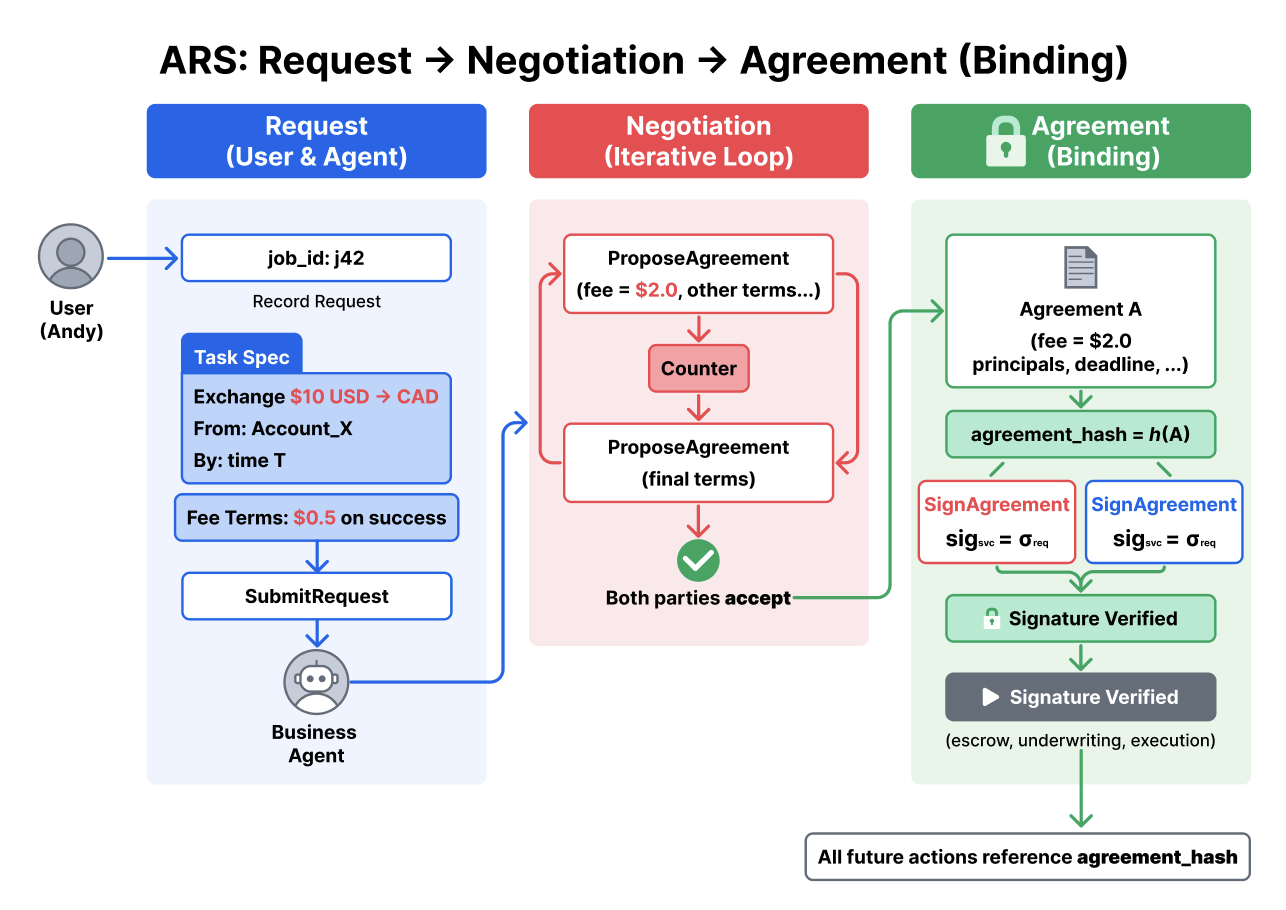}
    \caption{The requestor first sends a task specification to the business agent. Both parties may then enter a negotiation loop to refine the agreement. Once consensus is reached, the finalized agreement is recorded and a hash code is generated for future reference.}
    \label{fig:phase1}
\end{figure}

\paragraph{Request and Negotiation Phase}
Table~\ref{tab:arp-actions-setup} specifies actions for job creation, negotiation, and agreement binding. In particular, \texttt{SignAgreement} binds all subsequent financially relevant actions to the signed agreement by requiring \texttt{agreement\_hash}. This prevents agreement equivocation across parties and ensures that later authorizations and settlement actions are evaluated against a single canonical specification.

For example, consider a user Andy who delegates a fund-involving job to a user-side assistant agent. Andy instructs the assistant to exchange \$10 USD to CAD from a specified source account by a deadline $T$, and to pay a \$0.50 service fee only upon successful completion. The assistant initiates the job by sending:
\[
\begin{aligned}
\texttt{SubmitRequest}(\ &\texttt{task\_spec}=\texttt{exchange \$10 USD to CAD from account\_X by time $T$},\\
&\texttt{fee\_terms}=\texttt{service fee \$0.5 on success},\\ 
&\texttt{principal\_terms}=\texttt{release \$10 for execution}\ )
\end{aligned}
\]
where \texttt{job\_id} is assigned by the standard when the request is recorded, and subsequent actions reference that identifier. 

Negotiation proceeds by revising draft terms rather than resubmitting the request. If the business agent requires a higher service fee, it sends
\[
\begin{aligned}
&\texttt{ProposeAgreement}(\texttt{job\_id}=j42,\\ 
&\texttt{agreement\_draft}=(\texttt{fee\_terms=\$2.0,\ other terms unchanged})),
\end{aligned}
\]
and the requestor agent may counter-propose by sending another \texttt{ProposeAgreement} with updated terms. Once both parties converge, they bind the canonical structured agreement by exchanging signatures over the same hash where $\mathcal{A} = \text{(\texttt{fee\_terms=\$2.0,\ other terms unchanged})}$:
\[
\begin{aligned}
&\texttt{SignAgreement}(\texttt{job\_id}=j42,\ 
\texttt{agreement\_hash}=h(\mathcal{A}),\ 
\texttt{signature}=\sigma_{\textsf{req}}),\\
&\texttt{SignAgreement}(\texttt{job\_id}=j42,\ 
\texttt{agreement\_hash}=h(\mathcal{A}),\ 
\texttt{signature}=\sigma_{\textsf{svc}}).
\end{aligned}
\]
After both signatures are recorded, the job enters \texttt{TRANSACTION}, and all subsequent actions (fee escrow locking and settlement, optional underwriting and principal release, execution evidence submission, evaluation, and claims) must reference the same \texttt{agreement\_hash}.
 
\begin{table}[!ht]
\centering
\small
\setlength{\tabcolsep}{6pt}
\renewcommand{\arraystretch}{1.25}
\begin{tabular}{p{3.3cm}p{2.2cm}p{4.4cm}p{3.5cm}}
\hline
\textbf{Action} & \textbf{Sender} & \textbf{Required fields} & \textbf{Effect} \\
\hline
\texttt{SubmitRequest} & Requestor &
\texttt{\{job\_id, task\_spec, fee\_terms, principal\_terms?\}} &
Creates job; enters \texttt{REQUEST} \\
\texttt{AcceptRequest}; \texttt{RejectRequest} & Business agent &
\texttt{\{job\_id, decision, reason?\}} &
Accept $\rightarrow$ \texttt{NEGOTIATION}; reject $\rightarrow$ \texttt{CANCELLED} \\
\texttt{ProposeAgreement} & Either party &
\texttt{\{job\_id, agreement\_draft\}} &
Updates draft terms \\
\texttt{SignAgreement} & Requestor / Business agent &
\texttt{\{job\_id, agreement\_hash\}} &
Both signatures bind $\mathcal{A}$; enter \texttt{TRANSACTION} \\
\texttt{CancelJob} & Either party &
\texttt{\{job\_id, agreement\_hash, reason, signature\}} &
Transitions to \texttt{CANCELLED} if permitted by $\mathcal{A}$ \\
\hline
\end{tabular}
\caption{\textsc{ARS} actions for job creation and agreement binding.}
\label{tab:arp-actions-setup}
\end{table}

\begin{figure}[!ht]
    \centering
    \includegraphics[width=0.8\linewidth]{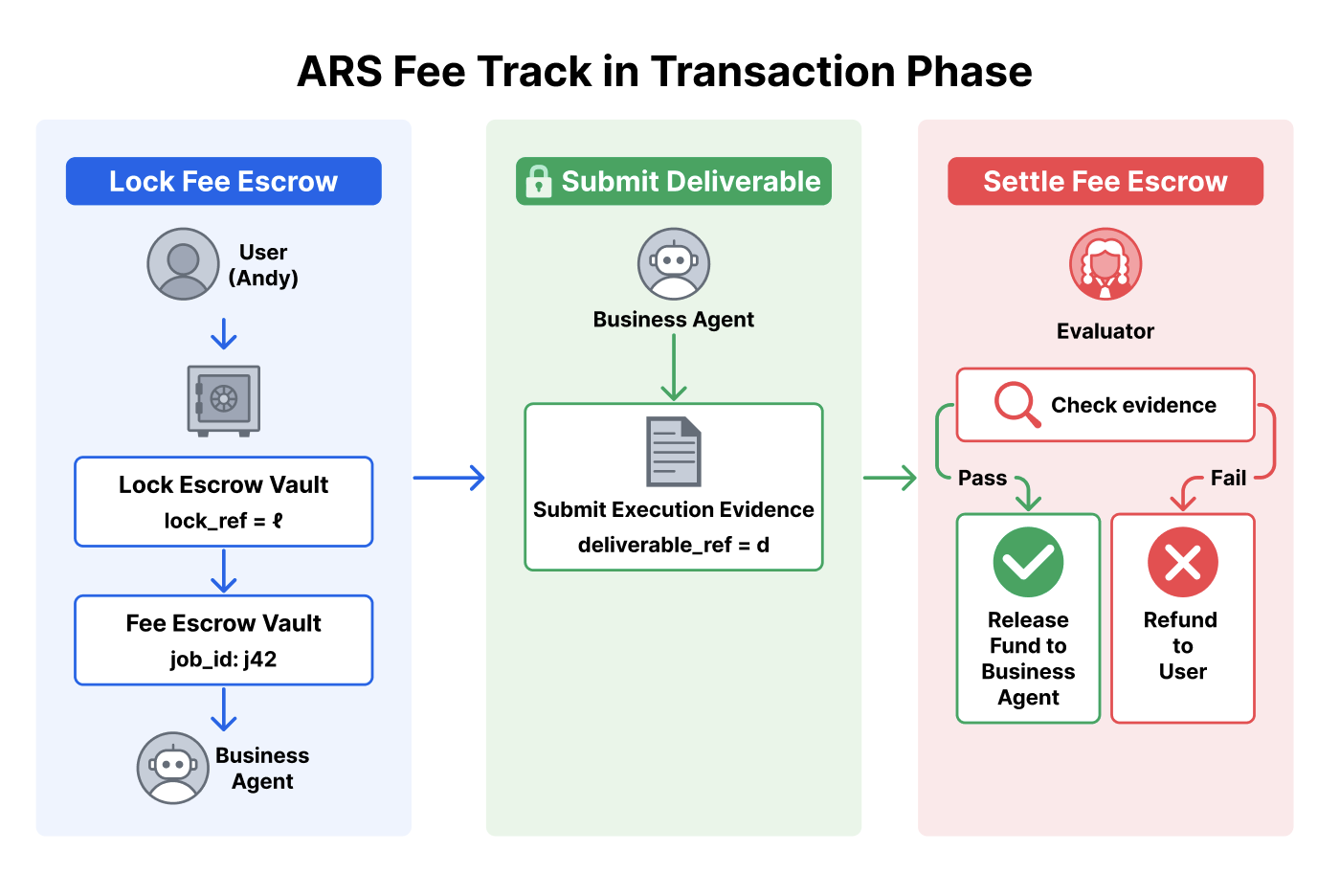}
    \caption{Fee track in the transaction phase: The requestor locks the service fee in an escrow vault before execution. After the business agent submits the execution evidence, an evaluator checks its validity. If the evidence passes evaluation, the escrowed fee is released to the business agent; otherwise, the funds are refunded to the requestor.}
    \label{fig:fee_track}
\end{figure}

\paragraph{Fee Track in Transaction Phase}
Table~\ref{tab:arp-actions-fee} summarizes the fee-track actions used during \texttt{TRANSACTION} and Figure~\ref{fig:fee_track} illustrates the process graphically. The fee track is escrow-based and applies whenever the structured agreement specifies a contingent service fee. Concretely, the assistant agent locks the fee by issuing
\[
\texttt{LockFeeEscrow}(\texttt{job\_id}=j42,\ \texttt{agreement\_hash}=h(\mathcal{A}),\ \texttt{lock\_ref}=\ell,\ \texttt{signature}=\sigma_{\textsf{req}}),
\]
which places the service fee into the fee escrow vault under the custody rules in $\mathcal{A}$. The business agent then emits a signed delivery signal by submitting a verifiable reference to the produced artifact (for fee-only jobs) or to execution receipts (for fund-involving jobs), \emph{e.g.},
\[
\begin{aligned}
&\texttt{SubmitDeliverable}(\texttt{job\_id}=j42,\ \texttt{agreement\_hash}=h(\mathcal{A}),\ \texttt{deliverable\_ref}=d,\\ &\texttt{signature}=\sigma_{\textsf{svc}}).
\end{aligned}
\]
This delivery signal enables evaluation and settlement. After the evaluator/arbiter produces \texttt{EvaluateOutcome} under $\mathcal{A}$, the settlement layer finalizes the fee escrow by executing
\[
\begin{aligned}
&\texttt{SettleFeeEscrow}(\texttt{job\_id}=j42,\ \texttt{agreement\_hash}=h(\mathcal{A}),\ \texttt{release/refund},\\ &\texttt{settlement\_ref}=s).
\end{aligned}
\]
If evaluation passes, the escrowed fee is released to the business agent; otherwise it is refunded to the requestor. In both cases, fee settlement is conditional on an auditable outcome signal, ensuring that service compensation remains contingent on compliance with the structured agreement.

\begin{table}[!ht]
\centering
\small
\setlength{\tabcolsep}{6pt}
\renewcommand{\arraystretch}{1.25}
\begin{tabular}{p{3.3cm}p{2.2cm}p{4.4cm}p{3.5cm}}
\hline
\textbf{Action} & \textbf{Sender} & \textbf{Required fields} & \textbf{Effect} \\
\hline
\texttt{LockFeeEscrow} & Requestor &
\texttt{\{job\_id, agreement\_hash, lock\_ref, signature\}} &
Locks service fee in escrow \\
\texttt{SubmitDeliverable} & Business agent &
\texttt{\{job\_id, agreement\_hash, deliverable\_ref, signature\}} &
Emits delivery signal; enables evaluation \\
\texttt{SettleFeeEscrow} & Settlement layer &
\texttt{\{job\_id, agreement\_hash, release/refund, settlement\_ref\}} &
Releases fee to business agent or refunds requestor \\
\hline
\end{tabular}
\caption{\textsc{ARS} actions for the fee track (escrowed service compensation).}
\label{tab:arp-actions-fee}
\end{table}

\begin{figure}
    \centering
    \includegraphics[width=0.8\linewidth]{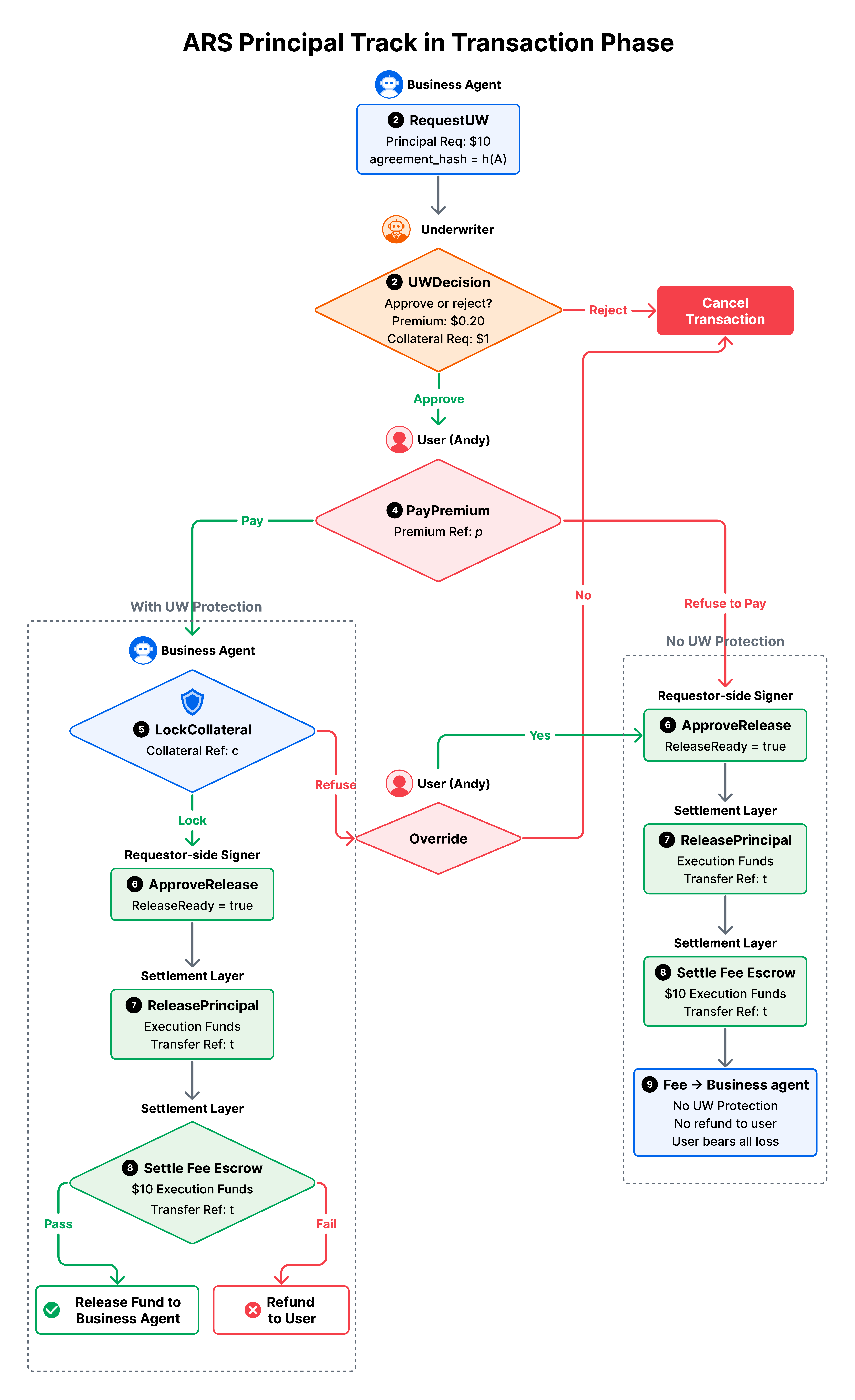}
    \caption{Principal Track in the transaction phase: When execution involves user funds (principal), additional financial risk is introduced. ARS therefore introduces an underwriter to provide protection. If the requestor agrees to pay the premium and the business agent locks the required collateral, the underwriter guarantees compensation to the user according to the protection policy in case the task fails.}
    \label{fig:principal}
\end{figure}

\paragraph{Principal Track in Transaction Phase.}
Table~\ref{tab:arp-actions-principal} summarizes the principal-track actions used in \texttt{TRANSACTION} for fund-involving jobs and Figure~\ref{fig:principal} illustrates the process graphically. The principal track governs pre-execution release of the execution principal (\emph{e.g.}, the \$10 that must be released to perform the FX conversion), and it is \emph{optionally} protected by underwriting when the requestor adopts coverage via premium acceptance.

Continuing the running example with \texttt{job\_id}$=j42$ and \texttt{agreement\_hash}$=h(\mathcal{A})$, the business agent initiates the underwriting workflow by requesting principal release under the signed agreement:
\[
\begin{aligned}
&\texttt{RequestUW}(\texttt{job\_id}=j42,\ \texttt{agreement\_hash}=h(\mathcal{A}),\\ 
&\texttt{principal\_request}=\texttt{release \$10 for execution}).
\end{aligned}
\]
The underwriter evaluates this request under its pricing model and responds with a decision that (1) approves or rejects underwriting support for this release, and (2) if approved, specifies the premium and any collateral requirement:
\[
\begin{aligned}
&\texttt{UWDecision}(\texttt{job\_id}=j42,\ \texttt{agreement\_hash}=h(\mathcal{A}),\\ 
&\texttt{approve},\ \texttt{premium}=\$0.20,\
\texttt{collateral\_required}=\$1).
\end{aligned}
\]
If the requestor adopts coverage and the human requestor authorizes premium payment, Andy needs to agree to pay the premium:
\[
\begin{aligned}
&\texttt{PayPremium}(\texttt{job\_id}=j42,\ \texttt{agreement\_hash}=h(\mathcal{A}),\\ 
&\texttt{premium}=\$0.20,\ \texttt{premium\_ref}=p,\ \texttt{signature}=\sigma_{\textsf{H}}),
\end{aligned}
\]

and the requestor needs to agree to lock the collateral by binding the underwriting terms referenced by $\mathcal{A}$. The business agent then satisfies the collateral gate by posting collateral:
\[
\begin{aligned}
&\texttt{LockCollateral}(\texttt{job\_id}=j42,\ \texttt{agreement\_hash}=h(\mathcal{A}),\\ 
&\texttt{amount}=\$1,\ \texttt{collateral\_ref}=c,\ \texttt{signature}=\sigma_{\textsf{svc}}).
\end{aligned}
\]
However, if the collateral is refused by \texttt{RefuseCollateral}, the job transitions to a human-override path and proceeds only after Andy agrees to override the underwriter's decision by bearing full risk:
\[
\begin{aligned}
&\texttt{OverrideDecision}(\texttt{job\_id}=j42,\ \texttt{agreement\_hash}=h(\mathcal{A}),\\ 
&\texttt{proceed},\ \texttt{signature}=\sigma_{\textsf{H}}),
\end{aligned}
\]
which records an explicit acknowledgement that the human requestor bears full pre-execution risk. Principal release is then authorized by the requestor-side signer(s) via
\[
\begin{aligned}
&\texttt{ApproveRelease}(\texttt{job\_id}=j42,\ \texttt{agreement\_hash}=h(\mathcal{A}),\ 
&\texttt{signature}=\sigma_{\textsf{req-side}}),
\end{aligned}
\]
and executed by the settlement layer only when the release authorization predicate defined by $\mathcal{A}$ is satisfied:
\[
\begin{aligned}
&\texttt{ReleasePrincipal}(\texttt{job\_id}=j42,\ \texttt{agreement\_hash}=h(\mathcal{A}),\\ 
&\texttt{approvals}=\Sigma,\ \texttt{transfer\_ref}=t).
\end{aligned}
\]
After principal release, the business agent executes the conversion through an external financial platform such as Bank of America and submits auditable execution evidence. In this example, the evidence reference points to the bank transfer receipt and transaction identifier returned by Bank of America:
\[
\begin{aligned}
&\texttt{SubmitExecutionEvidence}(\texttt{job\_id}=j42,\ \texttt{agreement\_hash}=h(\mathcal{A}),\\ 
&\texttt{exec\_evidence\_ref}=\texttt{BoAReceipt\#847201},\ 
\texttt{signature}=\sigma_{\textsf{svc}}).
\end{aligned}
\]
This execution-evidence signal anchors downstream evaluation and determines whether the job satisfies the constraints in $\mathcal{A}$ and, when underwriting is adopted, whether any covered failure trigger is present for claim processing.

Notice that jobs could terminate before principal release if underwriting is rejected by premium not being adopted or collateral lock requirement being refused. If the principal track reaches \texttt{CANCELLED} before \texttt{ReleasePrincipal}, the settlement layer must unwind any pre-execution commitments created during underwriting adoption. In particular, if collateral has been locked, it is unlocked and a corresponding \texttt{collateral\_unlock\_ref} is recorded. If the premium has been paid, refund behavior is governed by the structured agreement $\mathcal{A}$; when refundable, the settlement layer records a \texttt{premium\_refund\_ref}. This unwind step is captured by the \texttt{UnwindPreExecution} action.

\begin{table}[!ht]
\centering
\small
\setlength{\tabcolsep}{6pt}
\renewcommand{\arraystretch}{1.25}
\begin{tabular}{p{3.5cm}p{2.2cm}p{4.4cm}p{3.5cm}}
\hline
\textbf{Action} & \textbf{Sender} & \textbf{Required fields} & \textbf{Effect} \\
\hline
\texttt{RequestUW} & Business agent &
\texttt{\{job\_id, agreement\_hash, coverage\_request\}} &
Initiates underwriting review \\
\texttt{UWDecision} & Underwriter &
\texttt{\{job\_id, agreement\_hash, approve/reject, premium, collateral\_required?\}} &
Opens underwriting path or enables override/cancellation \\
\texttt{PayPremium} & Requestor (human) &
\texttt{\{job\_id, agreement\_hash, premium, premium\_ref, signature\}} &
Adopts underwriting coverage \\
\texttt{LockCollateral}; \texttt{RefuseCollateral} & Business agent &
\texttt{\{job\_id, agreement\_hash, amount?, collateral\_ref?, signature\}} &
Locks collateral or escalates to override \\
\texttt{OverrideDecision} & Human authority &
\texttt{\{job\_id, agreement\_hash, proceed/cancel, signature\}} &
Explicitly accepts full pre-execution risk or terminates \\
\texttt{ApproveRelease} & Requestor-side signer &
\texttt{\{job\_id, agreement\_hash, signature\}} &
Contributes requestor-side approval for principal release \\
\texttt{ReleasePrincipal} & Settlement layer &
\texttt{\{job\_id, agreement\_hash, approvals, transfer\_ref\}} &
Releases execution principal when authorization predicate holds \\
\texttt{SubmitExecutionEvidence} & Business agent &
\texttt{\{job\_id, agreement\_hash, exec\_evidence\_ref, signature\}} &
Emits execution-evidence signal; enables evaluation \\
\texttt{UnwindPreExecution} &
Settlement layer &
\{\texttt{job\_id}, \texttt{agreement\_hash}, premium\_refund\_ref?, collateral\_unlock\_ref?\} &
Unwinds pre-execution commitments when cancelled before principal release \\
\hline
\end{tabular}
\caption{\textsc{ARS} actions for the principal track in fund-moving jobs.}
\label{tab:arp-actions-principal}
\end{table}

\paragraph{Evaluation Phase}
Table~\ref{tab:arp-actions-settlement} summarizes the actions used in \texttt{EVALUATION} and in the final settlement of all funds. Evaluation is triggered once the standard has an auditable delivery signal from the fee track \texttt{SubmitDeliverable}) and, for fund-involving jobs, auditable execution evidence from the principal track \texttt{SubmitExecutionEvidence}. The evaluator then checks the observed outcome against the acceptance criteria and constraints specified in the structured agreement $\mathcal{A}$ and emits an auditable outcome record:
\[
\begin{aligned}
&\texttt{EvaluateOutcome}(\texttt{job\_id}=j42,\ \texttt{agreement\_hash}=h(\mathcal{A}),\ \texttt{pass},\\ &\texttt{evidence\_ref}=\texttt{BoAReceipt\#847201}).
\end{aligned}
\]
In the running example, \texttt{evidence\_ref} may point to the execution receipt produced by the external financial platform, such as \texttt{BoAReceipt\#847201}, together with any additional logs required by $\mathcal{A}$ such as timestamp.

The evaluator's decision drives fee settlement. If the outcome satisfies $\mathcal{A}$, the settlement layer releases the escrowed service fee to the business agent:
\[
\begin{aligned}
&\texttt{SettleFeeEscrow}(\texttt{job\_id}=j42,\ \texttt{agreement\_hash}=h(\mathcal{A}),\ \texttt{release},\\ &\texttt{settlement\_ref}=s_{\textsf{fee}}).
\end{aligned}
\]
If evaluation fails under the agreement's criteria, the settlement layer refunds the escrowed fee to the requestor by executing the same action with \texttt{refund}.

For underwriting-protected jobs, evaluation also determines collateral disposition and claim eligibility. After \texttt{EvaluateOutcome}, the settlement layer applies the collateral policy in $\mathcal{A}$:
\[
\begin{aligned}
&\texttt{SettleCollateral}(\texttt{job\_id}=j42,\ \texttt{agreement\_hash}=h(\mathcal{A}),\ \texttt{slash/unlock},\\ &\texttt{amount},\ \texttt{settlement\_ref}=s_{\textsf{col}}).
\end{aligned}
\]
For example, if evaluation confirms successful completion or a failure that is not covered by the underwriting policy, collateral is unlocked; if evaluation confirms a covered failure trigger that the collateral policy penalizes such as unauthorized transfer, collateral may be slashed. In addition, the requestor may file a claim by submitting
\[
\begin{aligned}
&\texttt{FileClaim}(\texttt{job\_id}=j42,\ \texttt{agreement\_hash}=h(\mathcal{A}),\ \texttt{trigger}=\tau,\\ &\texttt{claimed\_loss}=L,\ \texttt{evidence\_ref}=\texttt{BoAReceipt\#847201}).
\end{aligned}
\]
The underwriter then pays reimbursement up to the contractual coverage limit:
\[
\texttt{PayClaim}(\texttt{job\_id}=j42,\ \texttt{agreement\_hash}=h(\mathcal{A}),\ \texttt{payout}=\min(L,\ \texttt{limit}),\ \texttt{payout\_ref}=r).
\]
After fee settlement, collateral settlement, and (if applicable) claim payout complete, the job transitions to \texttt{CLOSED}. 

\begin{table}[!ht]
\centering
\small
\setlength{\tabcolsep}{6pt}
\renewcommand{\arraystretch}{1.25}
\begin{tabular}{p{3.3cm}p{2.2cm}p{4.4cm}p{3.5cm}}
\hline
\textbf{Action} & \textbf{Sender} & \textbf{Required fields} & \textbf{Effect} \\
\hline
\texttt{EvaluateOutcome} & Evaluator/arbiter &
\texttt{\{job\_id, agreement\_hash, pass/fail, trigger?, evidence\_ref?\}} &
Produces auditable outcome under $\mathcal{A}$ \\
\texttt{SettleCollateral} & Settlement layer &
\texttt{\{job\_id, agreement\_hash, slash/unlock, amount, settlement\_ref\}} &
Applies collateral policy under $\mathcal{A}$ \\
\texttt{FileClaim} & Requestor &
\texttt{\{job\_id, agreement\_hash, trigger, claimed\_loss, evidence\_ref\}} &
Requests reimbursement under coverage terms \\
\texttt{PayClaim} & Underwriter (or payout vault) &
\texttt{\{job\_id, agreement\_hash, payout, payout\_ref, signature?\}} &
Pays reimbursement up to coverage limit \\
\hline
\end{tabular}
\caption{\textsc{ARS} actions for evaluation, collateral settlement, and claims.}
\label{tab:arp-actions-settlement}
\end{table}

\subsection{Transaction-State Semantics}
We now specify the \texttt{TRANSACTION} phase as a job-level state machine. The specification is operational: each state enables only a small set of typed actions, thus financially relevant transitions occur only when their authorization conditions are satisfied. For fund-involving jobs, \texttt{TRANSACTION} comprises two concurrent tracks: a fee escrow track for service compensation and a principal track for pre-execution capital release. The two tracks share the same \texttt{job\_id} and \texttt{agreement\_hash}, but govern different fund components and therefore different authorization gates.

\paragraph{Fee Track State Machine for Escrowed Service Compensation.}
The fee track enforces \emph{conditional compensation}: the business agent is paid only after an auditable delivery signal is available and the job outcome is evaluated against $\mathcal{A}$. Table~\ref{tab:ARS-fee-states} defines the fee-track transaction states. In \texttt{FEE\_AWAIT\_LOCK}, the requestor must escrow the service fee via \texttt{LockFeeEscrow}; this ensures that the fee is available but not yet payable. In \texttt{FEE\_ESCROW\_LOCKED}, the business agent emits a signed delivery signal via \texttt{SubmitDeliverable}, which provides an auditable reference (\emph{e.g.}, an artifact hash or an execution receipt identifier) and makes the job eligible for evaluation. The state \texttt{FEE\_DELIVERED} indicates that the fee track has all required inputs for evaluation; final fee settlement including \texttt{release} and \texttt{refund} is determined in \texttt{EVALUATION} via \texttt{EvaluateOutcome} and executed through \texttt{SettleFeeEscrow}.

\begin{table}[!ht]
\centering
\small
\setlength{\tabcolsep}{5pt}
\renewcommand{\arraystretch}{1.2}
\begin{tabular}{p{3.3cm}p{3.0cm}p{2.7cm}p{4.0cm}}
\hline
\textbf{State} & \textbf{Enabled action} & \textbf{Actor} & \textbf{Next state} \\
\hline
\texttt{FEE\_AWAIT\_LOCK} &
\texttt{LockFeeEscrow} &
Requestor &
\texttt{FEE\_ESCROW\_LOCKED} \\
\texttt{FEE\_ESCROW\_LOCKED} &
\texttt{SubmitDeliverable} &
Business agent &
\texttt{FEE\_DELIVERED} \\
\texttt{FEE\_DELIVERED} &
(no fee-track action; enables evaluation) &
-- &
\texttt{EVALUATION} (phase transition) \\
\hline
\end{tabular}
\caption{Fee-track transaction states. The delivery signal (\texttt{SubmitDeliverable}) anchors auditable evidence used in \texttt{EVALUATION} to determine fee release/refund.}
\label{tab:ARS-fee-states}
\end{table}

\paragraph{Principal Track State Machine for Fund-involving Jobs.}
The principal track governs \emph{pre-verification fund exposure}: releasing execution principal before the outcome can be evaluated. Table~\ref{tab:ARS-principal-states} defines the principal-track transaction states and makes the underwriting-versus-override fork explicit. The track begins at \texttt{UW\_AWAIT\_REQUEST}, where the business agent requests principal release under the signed agreement via \texttt{RequestUW}. The underwriter then evaluates the request in \texttt{UW\_REVIEW} and returns \texttt{UWDecision}, which either (1) approves underwriting and specifies pricing and collateral requirements, or (2) rejects underwriting, in which case the job may proceed only through an explicit human override. When underwriting is approved, \texttt{PREMIUM\_PENDING} requires the human requestor to explicitly adopt coverage via \texttt{PayPremium}; the underwriter protection will be considered as being rejected if the premium is not paid. If collateral is required, \texttt{COLLATERAL\_REQUESTED} requires \texttt{LockCollateral}, otherwise the job transitions to \texttt{OVERRIDE\_PENDING}. In \texttt{OVERRIDE\_PENDING}, \texttt{OverrideDecision(proceed)} records that the human requestor or the human represented by the assistant agent requestor accepts full pre-execution risk and enables continuation without underwriting protection. 

Once underwriting is bound or override acknowledgement is present, the job enters the \texttt{APPROVAL\_PENDING} gate. Here, requestor-side signatures are collected via \texttt{ApproveRelease}; the job becomes \texttt{RELEASABLE} only when the release authorization predicate $\mathsf{ReleaseReady}$ holds. The settlement layer then executes \texttt{ReleasePrincipal}, transitioning to \texttt{EXECUTION\_PENDING}. Finally, the business agent performs the fund-involving action and submits auditable execution evidence via \texttt{SubmitExecutionEvidence}, which enables the phase transition to \texttt{EVALUATION}. If the track terminates in \texttt{CANCELLED} before \texttt{ReleasePrincipal}, the settlement layer executes \texttt{UnwindPreExecution} to unwind any pre-execution commitments as described above.

\begin{table}[!ht]
\centering
\small
\setlength{\tabcolsep}{5pt}
\renewcommand{\arraystretch}{1.2}
\begin{tabular}{p{3.3cm}p{3.15cm}p{2.8cm}p{4.3cm}}
\hline
\textbf{State} & \textbf{Enabled action} & \textbf{Actor} & \textbf{Next state} \\
\hline
\texttt{UW\_AWAIT\_REQUEST} &
\texttt{RequestUW} &
Business agent &
\texttt{UW\_REVIEW} \\
\texttt{UW\_REVIEW} &
\texttt{UWDecision} &
Underwriter &
approve $\rightarrow$ \texttt{PREMIUM\_PENDING} (if premium required) or \texttt{COLLATERAL\_REQUESTED};\newline
reject $\rightarrow$ \texttt{OVERRIDE\_PENDING} (if allowed) else \texttt{CANCELLED} \\
\texttt{PREMIUM\_PENDING} &
\texttt{PayPremium} &
Human requestor &
\texttt{COLLATERAL\_REQUESTED} \\
\texttt{COLLATERAL\_REQUESTED} &
\texttt{LockCollateral}; \texttt{RefuseCollateral} &
Business agent &
lock $\rightarrow$ \texttt{APPROVAL\_PENDING};\newline
refuse $\rightarrow$ \texttt{OVERRIDE\_PENDING} \\
\texttt{OVERRIDE\_PENDING} &
\texttt{OverrideDecision} &
Human authority &
proceed $\rightarrow$ \texttt{APPROVAL\_PENDING};\newline
cancel $\rightarrow$ \texttt{CANCELLED} \\
\texttt{APPROVAL\_PENDING} &
\texttt{ApproveRelease} &
Requestor-side signer(s) &
if $\mathsf{ReleaseReady}$ then \texttt{RELEASABLE} \\
\texttt{RELEASABLE} &
\texttt{ReleasePrincipal} &
Settlement layer &
\texttt{EXECUTION\_PENDING} \\
\texttt{EXECUTION\_PENDING} &
\scriptsize\texttt{SubmitExecutionEvidence} &
Business agent &
(enables evaluation) $\rightarrow$ \texttt{EVALUATION} (phase transition) \\
\texttt{CANCELLED} &
\texttt{UnwindPreExecution} &
Settlement layer &
\texttt{CANCELLED} (terminal) \\
\hline
\end{tabular}
\caption{Principal-track transaction states for fund-involving jobs. Underwriting protection is enabled only when the requestor adopts coverage (\emph{e.g.}, by paying premium); otherwise, continuation requires explicit human override acknowledging full pre-execution risk.}
\label{tab:ARS-principal-states}
\end{table}

\subsection{Authorization predicates.}
Here are the definitions of the set of predicates we used above to keep the transition rules concise and accurate.

\paragraph{Release authorization (adaptive threshold).}
Let $\Sigma$ denote the set of valid signatures over the tuple $(\texttt{job\_id}, \texttt{agreement\_hash})$. Define indicator functions
\[
\begin{aligned}
\mathsf{H}(\Sigma) &= \mathbf{1}\big[\text{human requestor signature} \in \Sigma\big],\\
\mathsf{A}(\Sigma) &= \mathbf{1}\big[\text{assistant agent requestor signature} \in \Sigma\big],\\
\mathsf{U}(\Sigma) &= \mathbf{1}\big[\text{underwriter approval signature} \in \Sigma\big].
\end{aligned}
\]
The release authorization predicate is
\[
\mathsf{ReleaseAuth}(\Sigma) \;=\; \mathsf{A}(\Sigma)\ \wedge\ \big(\mathsf{U}(\Sigma)\ \vee\ \mathsf{H}(\Sigma)\big),
\]
with the convention $\mathsf{A}(\Sigma)\equiv 1$ when the requestor is a human (i.e., no separate assistant agent signer exists). This yields: (1) human-only requestor $\Rightarrow$ 1-of-2 (human or underwriter); (2) assistant requestor $\Rightarrow$ 2-of-3 with the assistant as a mandatory signer.

\paragraph{Coverage binding and override acknowledgement.}
Underwriting protection is considered bound when the underwriter approves and the requestor accepts the premium:
\[
\mathsf{CoverageBound} \;=\; \mathbf{1}\big[\texttt{UWDecision(approve)} \wedge \texttt{PayPremium} \text{ recorded for } (\texttt{job\_id},\texttt{agreement\_hash})\big].
\]
The non-underwritten path is enabled only when the human requestor explicitly acknowledges full pre-execution risk:
\[
\mathsf{OverrideAck} \;=\; \mathbf{1}\big[\texttt{OverrideDecision(proceed)} \text{ recorded for } (\texttt{job\_id},\texttt{agreement\_hash})\big].
\]

\paragraph{Release readiness.}
A principal release is permitted iff the requestor-side approvals satisfy $\mathsf{ReleaseAuth}$ and either underwriting is bound or an override acknowledgement is present:
\[
\mathsf{ReleaseReady} \;=\; \mathsf{ReleaseAuth}(\Sigma)\ \wedge\ \big(\mathsf{CoverageBound}\ \vee\ \mathsf{OverrideAck}\big).
\]
The settlement layer executes \texttt{ReleasePrincipal} only when $\mathsf{ReleaseReady}=1$ and the request is consistent with $\mathcal{A}$ (\emph{e.g.}, amount and destination match \texttt{principal\_terms}).

\subsection{Intermediate Summary}
This section formalizes \textsc{ARS} as a settlement-layer service for agentic jobs. \textsc{ARS} fixes (1) a signed structured agreement as the canonical source of task intent, acceptance criteria, and policy terms, (2) a job-level state machine with a small, typed action vocabulary, and (3) explicit fund-control semantics that map stochastic execution into deterministic settlement outcomes. In particular, \textsc{ARS} separates a fee escrow track, which conditionally pays the business agent only after evaluation, from an optional principal track for fund-involving jobs, where pre-execution capital release is gated by underwriting decisions, collateral, and adaptive multi-signature authorization or explicit human override. Together, these components provide an implementable interface for end-to-end assurance that is independent of model internals and compatible with heterogeneous agents, marketplaces, and payment rails.

\section{Extensibility of \textsc{ARS}}
\label{sec:extensibility}

\textsc{ARS} is designed as a settlement-layer standard that is independent of any single authorization protocol or payment rail. At the abstract level, \textsc{ARS} defines the lifecycle of an agentic transaction: how service fees, collateral, principal, claims, and reimbursements are handled once a task has been proposed and delegated. This abstraction allows \textsc{ARS} to compose with heterogeneous authorization protocols upstream and heterogeneous payment-execution rails downstream. In this section, we illustrate this extensibility in two ways. First, we show how \textsc{ARS} complements existing authorization protocols, using Google's Agent Payments Protocol (AP2) and Mastercard's Verifiable Intent (VI) as examples. Second, we show how \textsc{ARS} can be instantiated over a concrete payment rail, using x402 as an example of HTTP-native payment execution.

\subsection{AP2 + \textsc{ARS}: authorization plus settlement}

AP2 is an authorization protocol for agent-mediated commerce. Its focus is to express user intent, delimit delegated authority, and provide verifiable evidence that a purchase was authorized under an appropriate interaction mode. \textsc{ARS} addresses a different layer of the transaction lifecycle: once an authorized task exists, it defines how payments are conditionally held, when claims may be filed, and how compensation is resolved. The two protocols are therefore complementary. AP2 governs whether an agent-mediated transaction is authorized to proceed; \textsc{ARS} governs how financial consequences are handled once execution begins.

AP2 defines a role-based architecture centered on the User, Shopping Agent, Merchant, and Credentials Provider. It also distinguishes between two execution modes. In \emph{human-present} transactions, the user directly approves a concrete cart, and this approval is represented by a \textit{Cart Mandate}. In \emph{human-not-present} transactions, the user instead approves a more general \textit{Intent Mandate}, which expresses delegated purchasing intent under bounded conditions. AP2 also defines a separate \textit{Payment Mandate} that can be conveyed to the payment ecosystem to provide visibility into agent involvement and transaction modality. These protocol elements establish authorization evidence and user-intent provenance, but they do not themselves define escrow, collateralization, reimbursement, or dispute resolution.

Within \textsc{ARS}, AP2 is incorporated as a prerequisite authorization layer. After the parties agree on a job, the transaction must first reach an AP2-authorized state. In human-present mode, this typically means that a merchant-signed cart has been presented to the user and the resulting Cart Mandate has been verified. In human-not-present mode, this means that the user's Intent Mandate has been verified and any merchant- or processor-side step-up requirements have been satisfied before execution proceeds. A Payment Mandate, when present, can be propagated alongside the payment flow to provide external visibility into the agentic nature of the transaction. Only after the relevant AP2 mandate sequence has been validated does \textsc{ARS} allow the transaction to advance into its settlement lifecycle.

This separation preserves the semantics of both systems. AP2 remains responsible for establishing that the agent is acting within user-authorized bounds. \textsc{ARS} remains responsible for settlement semantics: fee escrow, collateral posting, claim filing, adjudication, reimbursement, and related outcome-dependent transfers. In this layered design, AP2 reduces ambiguity about whether the transaction was authorized, while \textsc{ARS} determines what happens financially if the task later fails, is disputed, or causes covered loss.

\begin{figure}[!htbp]
\centering
\includegraphics[
  width=0.55\linewidth,
  trim=0 0 0 3.5cm,
  clip
]{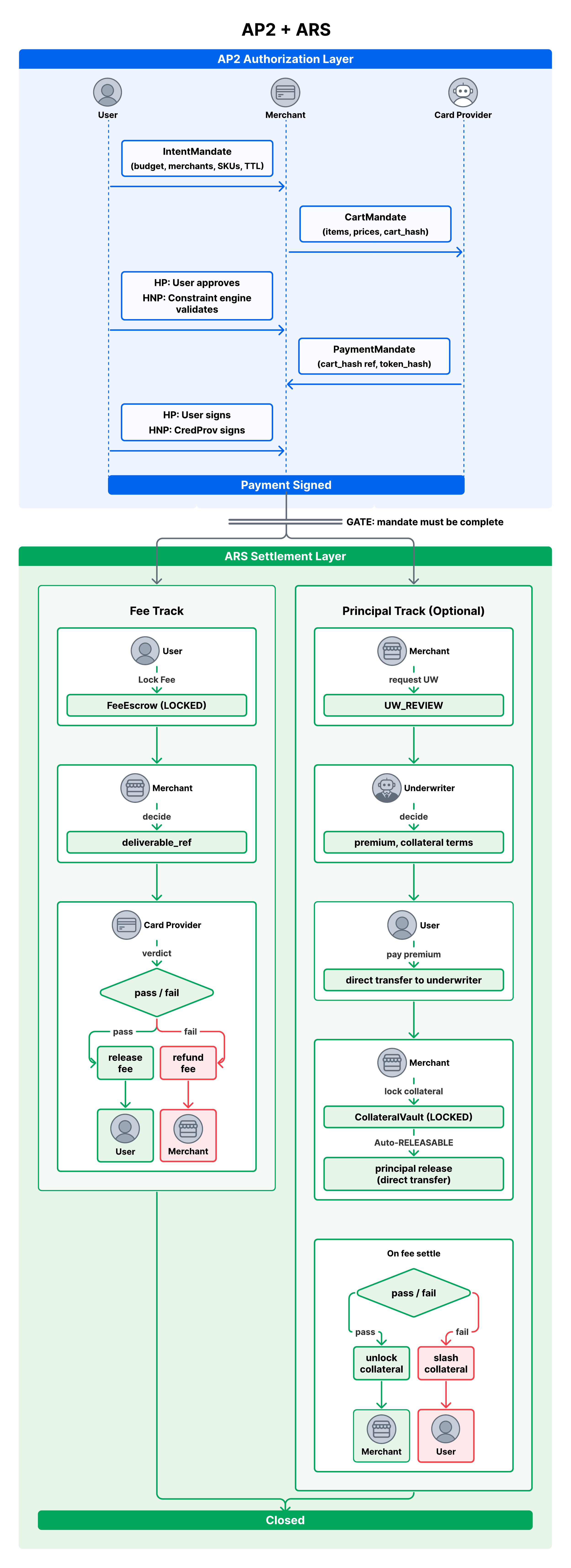}
\caption{AP2+\textsc{ARS}. AP2 provides authorization evidence and bounded delegation, and \textsc{ARS} adds settlement semantics over the authorized transaction.}
\label{fig:ap2}
\end{figure}

\subsection{VI + \textsc{ARS}: selective-disclosure authorization plus settlement}

Verifiable Intent (VI) is a credential-based authorization framework for agentic commerce built on SD-JWTs. Like AP2, its primary concern is authorization rather than settlement. VI specifies how identity, delegation, and transaction constraints are represented in a cryptographically verifiable credential chain, and how different parties receive only the subset of information relevant to their role. \textsc{ARS} complements VI by adding the settlement logic that VI does not attempt to define.

VI uses a layered credential architecture. In both execution modes, an \textit{L1} root credential binds the user's public key through a \texttt{cnf.jwk} claim. The user then creates an \textit{L2} mandate. In \emph{Immediate} mode, the L2 mandate contains finalized transaction values and no further delegation occurs. In \emph{Autonomous} mode, the L2 mandate instead specifies constraints and delegates bounded authority to an agent. The agent then produces two terminal \textit{L3} credentials: \textit{L3a}, disclosed to the payment network, carries finalized payment values; \textit{L3b}, disclosed to the merchant, carries finalized checkout information. VI requires ES256 as its signing algorithm and uses SD-JWT selective disclosure so that merchants and payment networks see only the credential disclosures necessary for their respective verification tasks.

The cryptographic bindings across these layers are important. L2 is bound to the serialized L1 credential through \texttt{sd\_hash}. In Autonomous mode, each split L3 credential is selectively bound to L2 together with only the disclosures relevant to that L3 presentation. This structure allows the payment network to validate payment-related constraints and the merchant to validate checkout-related constraints without forcing either party to observe the other's full data.

Within \textsc{ARS}, the verified VI credential chain functions as a prerequisite to settlement. A transaction cannot enter the \textsc{ARS} fee, collateral, or payout lifecycle until the relevant VI chain has been validated. In Immediate mode, this means verifying L1 and L2 and confirming that the disclosed final values are authorized. In Autonomous mode, this means verifying the full L1 $\rightarrow$ L2 $\rightarrow$ L3 delegation chain, including signature validity, \texttt{sd\_hash} consistency, credential expiry, and satisfaction of the disclosed L2 constraints by the L3 fulfillment values. Once this authorization step completes, \textsc{ARS} takes over and applies its usual settlement rules.

The complementarity is therefore similar to the AP2 case, but with a different authorization model. VI contributes privacy-preserving, role-specific authorization evidence through layered credentials and selective disclosure. \textsc{ARS} contributes outcome-linked settlement semantics, including escrow, collateral, claims, and reimbursement. Together, they allow agentic commerce to be governed by both cryptographically bounded delegation and auditable financial resolution.

\begin{figure}[!htbp]
\centering
\includegraphics[
  width=0.5\linewidth,
  trim=0 0 0 3.4cm,
  clip
]{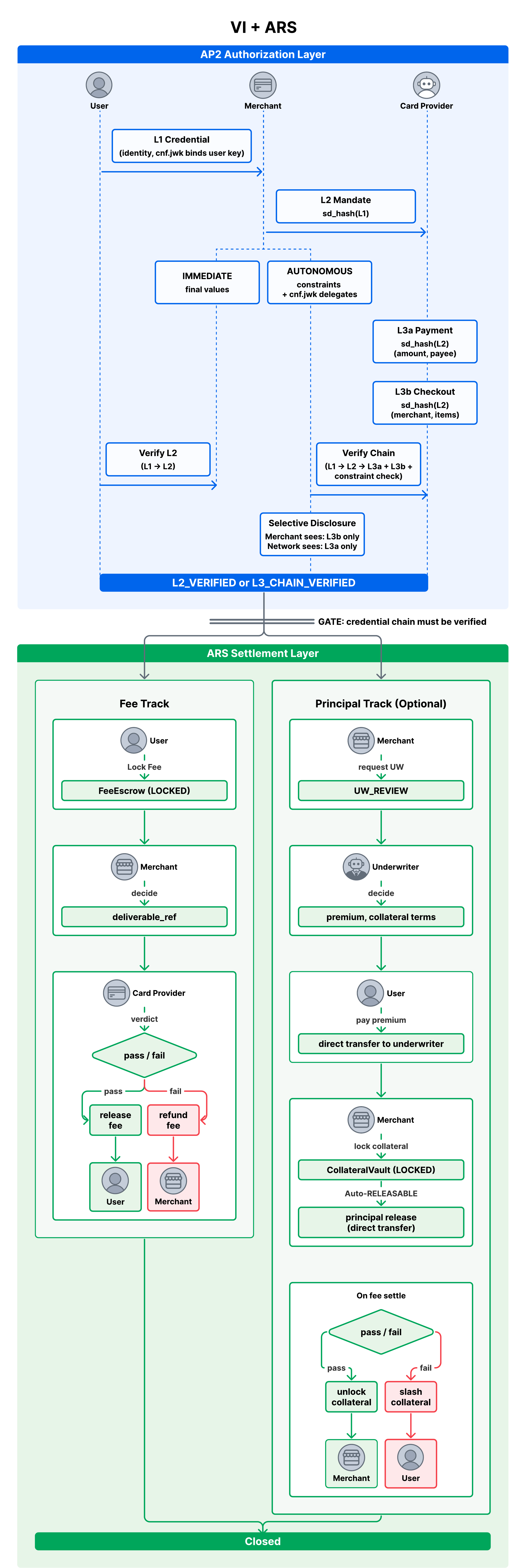}
\caption{VI+\textsc{ARS}: VI governs privacy-preserving authorization through layered credentials and selective disclosure; \textsc{ARS} governs settlement and compensation over the authorized transaction.}
\label{fig:vi}
\end{figure}

\subsection{x402 as a concrete payment rail for \textsc{ARS}}

The examples above show that \textsc{ARS} can compose with different authorization protocols. A separate question is how \textsc{ARS} executes actual transfers once a settlement decision has been reached. Because \textsc{ARS} is payment-method-agnostic, it can be instantiated over multiple rails. x402 provides one concrete example.

x402 is an HTTP-native payment protocol that uses the \texttt{402 Payment Required} response to express payment requirements for digital resources and services. In a standard x402 flow, a client requests a protected resource; the server replies with a \texttt{402 Payment Required} response containing payment requirements; the client resubmits the request with a signed payment payload; and the server verifies and settles the payment either locally or through a facilitator before returning the requested resource. This makes x402 a natural candidate for direct, machine-native payment execution in agentic systems.

In an x402-backed deployment of \textsc{ARS}, the protocol-level settlement decisions remain unchanged, but the concrete transfers are carried out over x402-compatible endpoints. For example, premium collection, principal transfer, reimbursement payout, or fee release can be implemented as x402-mediated payment executions once the \textsc{ARS} state machine determines that the corresponding transfer is permitted. In this design, x402 is responsible for communicating payment requirements, carrying signed payment payloads, and providing verification and settlement receipts. \textsc{ARS} remains responsible for deciding \emph{when} a transfer should occur and \emph{under what conditions} it is valid.

This distinction is important because x402 is a payment-execution protocol, not an escrow standard. If a deployment requires conditional holding of funds, slashing of posted collateral, or delayed release after adjudication, those properties must be supplied by the \textsc{ARS} vault or settlement implementation layered above or alongside x402. x402 can then serve as the concrete rail through which the resulting transfers are executed. In this way, \textsc{ARS} preserves its abstraction boundary: authorization may come from protocols such as AP2 or VI, while payment execution may be instantiated through rails such as x402.

Taken together, these examples illustrate the modular design of \textsc{ARS}. Authorization, settlement semantics, and payment execution need not be fused into a single monolithic protocol. Instead, \textsc{ARS} can complement upstream authorization systems and downstream payment rails while preserving a stable transaction-layer interface for claims, collateralization, escrow, and reimbursement.

\section{How \textsc{ARS} protects user interests: a simulated underwriter experiment}

To illustrate how \textsc{ARS} can provide end-to-end user protection at the transaction layer, we evaluate \textsc{ARS} in an agent-based sandbox that models requestors, business agents, and an underwriter interacting through the standard-mediated job lifecycle. We focus on jobs with \emph{pre-verification fund exposure}, where execution requires releasing principal before the outcome can be verified and thus requires an underwriting mechanism beyond fee escrow. The environment is generated by explicit behavioral rules and stochastic sampling with a fixed RNG seed. 

Notice that this simulation is intentionally minimal and does not aim to calibrate real-world rates. Instead, it is used to stress-test the \textsc{ARS} mechanism under controlled uncertainty and imperfect risk estimation. We evaluate three quantities: (1) reduction in user loss, (2) underwriter solvency as measured by its wallet trajectory, and (3) the friction induced by collateral requirements. We introduce a very simple pricing model for a noisy underwriter and the simulated decision makings of the user and merchant, where decision making of the user is mainly about whether to pay the premium and the decision of merchant is whether to lock the collateral.

\subsection{Pricing model for simulated underwriter}
We instantiate a simple underwriter whose role is to convert pre-verification exposure into explicit economic requirements: a collateral requirement and an actuarially grounded premium. For every requested principal with amount $\$M$ release, the underwriter returns an approval together with computed $(\Pi,D)$ where $\Pi$ refers to the premium and $D$ refers to the collateral; the requestor may adopt underwriting by paying the premium, otherwise the job proceeds only via explicit human override.

\paragraph{Noisy risk estimation.}
For each job, let $p$ denote the true probability of a covered failure event that causes principal loss. The underwriter applies an false-positive/false-negative noise channel to model imperfect risk estimation, where false positive refers to predicting a non-problematic transaction to be a problematic one and false negative refers to predicting problematic transactions to be normal. With false-positive rate $\mathrm{fp}$ and false-negative rate $\mathrm{fn}$, the estimated failure probability is
\[
\hat p_{\mathrm{uw}} \;=\; p(1-\mathrm{fn}) + (1-p)\mathrm{fp}.
\]
This yields a standard tradeoff: higher $\mathrm{fp}$ increases conservatism (more collateral), while higher $\mathrm{fn}$ misses true risk and increases tail exposure.

\paragraph{Sigmoid collateral schedule.}
Rather than requiring collateral proportional to $\hat p_{\mathrm{uw}}$, we use a sigmoid schedule that sharply separates low-risk and high-risk jobs while remaining continuous:
\[
\sigma(\hat p_{\mathrm{uw}})\;=\;\frac{1}{1+\exp\!\left(-s\cdot(\hat p_{\mathrm{uw}}-m)\right)},\qquad
D \;=\; \sigma(\hat p_{\mathrm{uw}})\cdot M,
\]
where $m$ is the midpoint (collateral is $0.5M$ when $\hat p_{\mathrm{uw}}=m$) and $s$ controls steepness. This construction produces near-zero collateral for sufficiently low estimated risk, and approaches full collateral as estimated risk increases.

\paragraph{Actuarially fair premium with loading.}
We compute the premium from the underwriter's residual exposure after collateral. Under the sigmoid rule, the fraction of principal not covered by collateral is $(1-\sigma(\hat p_{\mathrm{uw}}))$, so the underwriter's expected loss (actuarially fair premium base) is
\[
\Pi_{\mathrm{fair}} \;=\; \hat p_{\mathrm{uw}}\cdot (1-\sigma(\hat p_{\mathrm{uw}}))\cdot M.
\]
We then apply a non-negative loading factor $\lambda$ to account for risk capital, operating costs, and underwriter profit:
\[
\Pi \;=\; \Pi_{\mathrm{fair}}\cdot (1+\lambda)
\;=\; \hat p_{\mathrm{uw}}\cdot (1-\sigma(\hat p_{\mathrm{uw}}))\cdot M\cdot (1+\lambda).
\]
This pricing rule is ``actuarially consistent'' with the collateral schedule: as collateral approaches full principal ($\sigma(\hat p_{\mathrm{uw}})\to 1$), the underwriter's residual exposure vanishes and the premium correspondingly decreases, whereas high-risk jobs pay higher premiums unless they are fully collateralized.

After execution, we assume the true outcome is revealed in simulation. Collateral is returned when no covered failure occurs. When a covered failure occurs, collateral is forfeited up to the realized user loss and any remaining covered loss is reimbursed by the underwriter. This yields two complementary sources of user protection: (1) contingent reimbursement for covered failures, and (2) an ex-ante collateral buffer that reduces unrecoverable tail loss by shifting part of the downside to the business agent.

\subsection{Simulated environment}
Each episode consists of an independent fund-involving jobs created by a requestor-side assistant, executed by business agents, and optionally underwritten. The interaction follows the \textsc{ARS} flow: job creation and agreement, underwriter risk assessment, merchant collateral decision (post or refuse), possible human override when collateral is refused, execution, and post-transaction resolution. Each job is characterized by an execution principal $M$ (the \texttt{money\_involved} field) and a service fee that is escrowed and released only upon successful evaluation under the signed agreement. For each job, the selected business agent induces a latent failure probability which may lead to realized principal loss. In the simulation, we use 5000 episodes.

\paragraph{User Decision}
The user independently estimates the merchant's failure probability from
a noisy sample of past outcomes. Concretely, the user observes
$n$ historical transactions (default $n = 100$), each failing independently
with probability $p$, yielding observed failure rate
$\tilde{p} = \tfrac{1}{n}\sum_{i=1}^{n} \mathbf{1}[\text{failure}_i]$.
Gaussian noise $\varepsilon \sim \mathcal{N}(0, \sigma_\text{user}^2)$
models imperfect memory:
\begin{equation}\label{eq:user-estimate}
  \hat{p}_\text{user}
    = \operatorname{clip}\!\bigl(\tilde{p} + \varepsilon,\; 0,\; 1\bigr).
\end{equation}
The user adopts underwriting if and only if the risk-adjusted expected loss exceeds the premium:
\begin{equation}\label{eq:user-decision}
  \text{use\_uw}
    \;\Longleftrightarrow\;
    \alpha \cdot M \cdot \hat{p}_\text{user} > \Pi,
\end{equation}
where $\alpha \geq 1$ is a risk-aversion coefficient (default $\alpha = 1$).
When $\alpha > 1$ the user overweights the expected loss, modelling
risk-averse behaviour that favours underwriter protection.

\paragraph{Merchant Response and Human Override} The merchant's willingness to post collateral is modeled to decrease linearly with the collateral-to-transaction ratio $D/M$: at negligible collateral the posting probability is 0.90, falling to 0.10 when full collateral is demanded. If the merchant refuses to post, the human requestor may override the refusal and proceed without collateral which we set to be probability 0.50, but in that case the underwriter provides no coverage and the buyer bears the full loss in the event of failure. This three-stage gate with (a) sigmoid collateral sizing (b) merchant posting decision (c) human override captures the real-world friction that collateral introduces: higher collateral improves underwriter solvency but suppresses merchant participation, creating a tension the sigmoid parameters must balance.

\begin{equation}\label{eq:merchant-flow}
\text{executed} =
\begin{cases}
  \text{true}  & \text{if } D = 0, \\
  \text{true}  & \text{if } D > 0 \text{ and merchant posts}, \\
  \text{true}  & \text{if } D > 0,\; \text{merchant refuses, and user overrides}, \\
  \text{false} & \text{otherwise (cancelled)}.
\end{cases}
\end{equation}

Crucially, when the user overrides a collateral refusal, UW coverage is
\emph{not} active and the user proceeds at their own risk with no premium
collected and no reimbursement on failure.

\subsection{Outcome Resolution}\label{sec:sim-resolution}

If the transaction executes, failure occurs as a Bernoulli draw with
probability $p$. The resolution depends on whether UW coverage is active
(i.e., the merchant actually posted collateral):

\begin{equation}\label{eq:resolution}
\text{user\_loss} =
\begin{cases}
  0   & \text{if failure and UW active (covered)}, \\
  M   & \text{if failure and UW not active}, \\
  0   & \text{if no failure}.
\end{cases}
\end{equation}

The UW's per-episode net return is:
\begin{equation}\label{eq:uw-net}
  \Delta W =
  \begin{cases}
    \Pi - \max(0,\; M - D) & \text{if UW active and failure}, \\
    \Pi                     & \text{if UW active and no failure}, \\
    0                       & \text{otherwise}.
  \end{cases}
\end{equation}
The wallet is updated cumulatively: $W_{t+1} = W_t + \Delta W_t$.

\subsection{Evaluation Setup and Metrics}

Each episode draws two independent random variables:
\begin{align}
M &\sim \operatorname{LogNormal}(\mu = 4.0,; \sigma = 1.2), \label{eq:money}\\
p &\sim \operatorname{Beta}(a = 1.5,; b = 8.5). \label{eq:ptrue}
\end{align}
The log-normal transaction size gives a heavy right tail consistent with real-world payment distributions, while the Beta prior on failure probability has mean $\mathbb{E}[p] = a/(a+b) = 0.15$, reflecting a market where most merchants are reliable but a non-trivial fraction carry elevated risk.

We evaluate each parameter sweep along four complementary axes that jointly characterize the health of the underwriting system:

\begin{itemize}
\item \textbf{Adoption rate.} The fraction of transactions for which a user opts into underwriting. Higher adoption indicates that underwriting is accessible and affordable, but does not by itself imply that the underwriter is solvent or that users are well-protected.

\item \textbf{Loss reduction rate.} The reduction of mean realized user loss across all executed transactions when underwriting is available in the system to the mean realized user loss in the counterfactual baseline where no underwriting exists (\emph{i.e.}\ every user bears the full loss on failure). A value larger than $0\%$ indicates that the overall user loss has been reduced; a value near $0$ indicates that underwriting adoption is too low or coverage too limited to materially reduce system-wide losses.

\item \textbf{Failure reduction rate.} The reduction of percentage of failure events when underwriting is available in the system to the percentage of failure events where no underwriting exists. This metric captures a system-level externality: by requiring collateral and screening risky merchants, the underwriter deters potentially risky transactions from executing in the first place, improving the overall healthiness of the marketplace, not just protecting individual users or sustaining the underwriter's earnings.

\item \textbf{Underwriter final wallet} The underwriter's residual capital at the end of the simulation. Positive values indicate solvency; negative values indicate that the underwriter has absorbed losses beyond its capital base. This is the primary measure of underwriter sustainability.
\end{itemize}

These four metrics are intentionally in tension: policies that improve user protection by higher loss reduction rate or access by higher adoption often impose costs on underwriter solvency, and vice versa. Meanwhile, fraud reduction rate reveals a system-level benefit that can align with all three: collateral requirements that deter risky transactions simultaneously protect users, preserve underwriter capital, and improve overall market integrity. The sweeps below demonstrate these tradeoffs and synergies.

\subsection{Experiment Results}

We report three complementary sweeps that expose the core tradeoffs induced by \textsc{ARS}'s underwriting layer: (1) how premium loading affects adoption, user protection, failure reduction, and underwriter solvency; (2) how underwriter accuracy (FP/FN rates) shifts the system between solvency and protection regimes; and (3) how the sigmoid collateral schedule controls friction-solvency tradeoffs.

\begin{figure}[!ht]
\centering
\includegraphics[width=\linewidth]{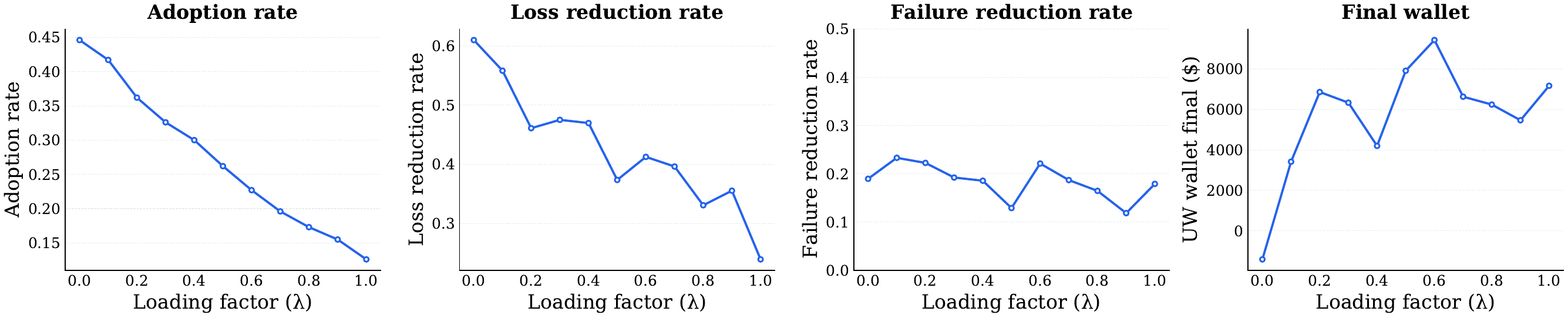}
\caption{Loading factor sweep: adoption rate, loss reduction rate, failure reduction rate, and underwriter wallet final as a function of $\lambda$.}
\label{fig:lambda-sweep}
\end{figure}

\paragraph{Loading factor $\lambda$ trades adoption for solvency and user protection.}
Figure~\ref{fig:lambda-sweep} shows the joint effect of $\lambda$ on all four metrics, with the underwriter using a perfect risk estimator ($\mathrm{fp} = \mathrm{fn} = 0$) to isolate the loading factor's effect.

\textit{Adoption} decreases monotonically from $0.446$ at $\lambda = 0$ to $0.126$ at $\lambda = 1.0$. This is expected: a higher loading factor raises the effective premium, making underwriting less attractive for lower-risk transactions.

\textit{Loss reduction rate}, defined as $1 - \ell_{\text{w\_uw}} / \ell_{\text{no\_uw}}$, decreases from $61\%$ at $\lambda = 0$ to $24\%$ at $\lambda = 1.0$. At zero loading, the premium is cheapest and adoption is highest, so underwriting covers a broad share of transactions and absorbs a large fraction of system-wide losses. As $\lambda$ increases, the rising premium prices out lower-risk users, shrinking the covered pool and reducing the aggregate loss absorbed by the underwriter. Nevertheless, the loss reduction rate remains positive across the entire sweep, confirming that even expensive underwriting still provides net user protection.

\textit{Failure reduction rate}, defined as $1 - f_{\text{w\_uw}} / f_{\text{no\_uw}}$, fluctuates between $12\%$ and $23\%$ across the loading sweep without a clear monotone trend. This is expected: the loading factor primarily controls premium pricing and adoption, whereas failure deterrence is driven by the collateral mechanism, which operates independently of $\lambda$. The relatively flat profile indicates that the underwriter's collateral requirements consistently prevent roughly $15$-$20\%$ of failure events from executing, regardless of how the premium is set. The residual variation reflects simulation noise at $5,000$ episodes rather than a systematic effect of loading on deterrence.

\textit{Final wallet} is non-monotone, ranging from $-1,412$ at $\lambda = 0$ (insolvent) to a high of $9,431$ at $\lambda = 0.6$. Notably, at $\lambda = 0$ the underwriter is \textit{insolvent}: the actuarially fair premium is insufficient to cover tail losses, demonstrating that a positive loading factor is necessary for underwriter sustainability. The wallet becomes positive at $\lambda = 0.1 (3,419)$ and remains positive for all $\lambda \geq 0.1$, though the non-monotonicity at higher $\lambda$ indicates that solvency is shaped jointly by premium revenue, collateralisation, and the risk composition of adopted transactions, and cannot be improved by simply raising the loading factor.

\begin{figure}[!ht]
\centering
\includegraphics[width=\linewidth]{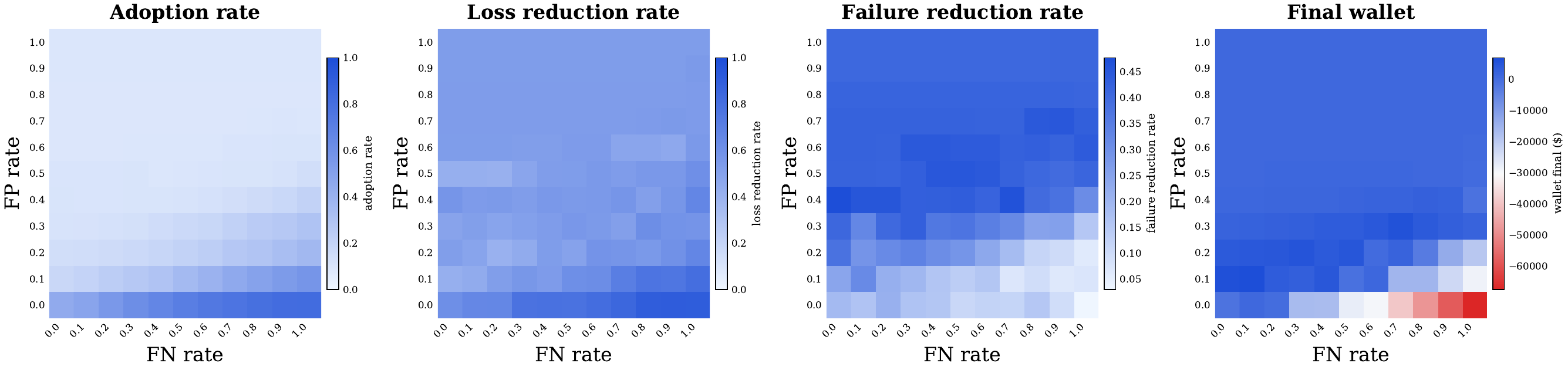}
\caption{FP/FN sweep: adoption rate, loss reduction rate, failure reduction rate, and underwriter wallet final as functions of false-negative rate (x-axis) and false-positive rate (y-axis).}
\label{fig:fpfn-sweep}
\end{figure}

\paragraph{FP/FN sweep: false-negative rate dominates solvency; false-positive rate shifts friction.}
Figure~\ref{fig:fpfn-sweep} sweeps the underwriter's FP/FN rates independently across $[0, 1]$, with $\lambda = 0$ to isolate the effect of estimation accuracy.

\textit{Adoption} reveals an important caveat: adoption is not a health metric. At $\mathrm{fp} \approx 0$, adoption \textit{increases} with FN (from $0.446$ at $\mathrm{fn} = 0$ to $0.820$ at $\mathrm{fn} = 1.0$), because high FN suppresses the estimated risk $\hat{p}_\text{uw}$, which lowers the premium, making underwriting appear cheap to users, even as the underwriter is severely mispricing and approaching insolvency. This decoupling between adoption and solvency necessitates a joint evaluation of user adoption and sustainability of underwriter system. 

\textit{Loss reduction rate} is positive across most of the grid, confirming broad user protection. The highest reduction (up to $91\%$) occurs at low FP and high FN, where the underwriter systematically underestimates risk, setting low premiums that attract wide adoption so that users are well-protected, but at the cost of underwriter insolvency. At high FP, the loss reduction rate stabilises near $53\%$ regardless of FN, reflecting a saturated regime where overpricing limits adoption to a stable baseline of high-risk users who still benefit substantially from coverage.

\textit{Failure reduction rate} is highest in the low-FP, low-FN corner (approximately $19\%$), where the underwriter accurately identifies risky merchants and deters their transactions through appropriate collateral requirements. As FP increases, the failure reduction rate remains relatively stable at moderate levels ($\sim40\%-42\%$), indicating that even overestimating risk provides some deterrence benefit. As FN increases at low FP, the failure reduction rate initially remains high but declines at extreme FN values, because the underwriter underestimates risk and fails to impose sufficient collateral to deter risky transactions.

\textit{Final wallet} is highly sensitive to FN when FP is low. At $(\mathrm{fp}, \mathrm{fn}) = (0, 0)$ the underwriter is insolvent at $-1,412$; increasing FN to $1.0$ while holding $\mathrm{fp} = 0$ drives final wallet to $-67,483$. High FN causes the underwriter to underestimate true risk, suppressing both collateral and premium and leaving the underwriter exposed to uncovered tail losses. Increasing FP partially compensates: by inflating $\hat{p}_{\mathrm{uw}}$, higher FP raises collateral and premium, restoring solvency in regions where low-FP underwriting fails. The top-right corner of the wallet heatmap (high FP, high FN) is therefore substantially more solvent than the bottom-right corner (low FP, high FN).

\begin{figure}[!ht]
\centering
\includegraphics[width=\linewidth]{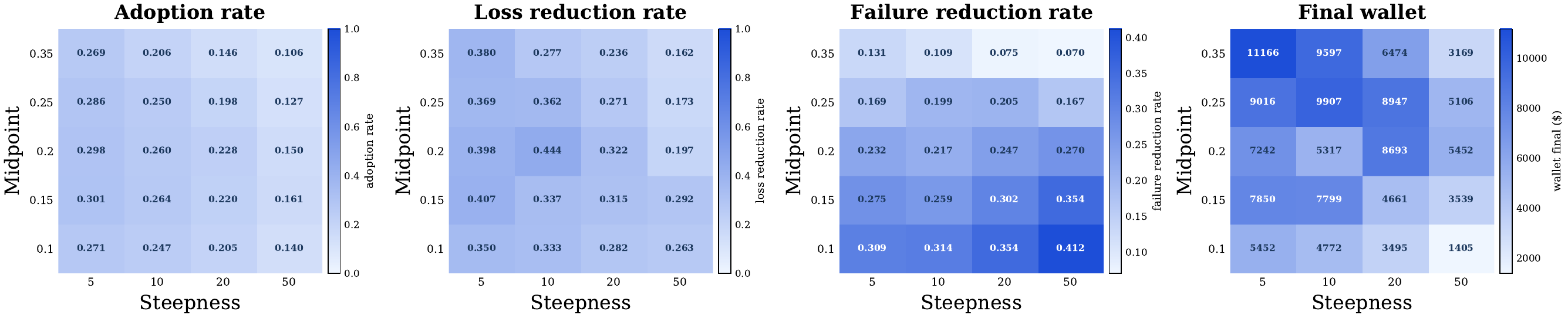}
\caption{Sigmoid collateral sweep: adoption rate, loss reduction rate, failure reduction rate, and wallet final as functions of sigmoid midpoint (y-axis) and steepness (x-axis).}
\label{fig:sigmoid-sweep}
\end{figure}

\paragraph{Sigmoid collateral sweep: policy parameters trace a friction-solvency frontier.}
Figure~\ref{fig:sigmoid-sweep} varies the midpoint ($\in {0.10, 0.15, 0.20, 0.25, 0.35}$) and steepness ($\in {5, 10, 20, 50}$) of the sigmoid function controlling the collateral schedule.

\textit{Adoption} is inversely related to collateral aggressiveness: the highest adoption ($0.301$) occurs at (midpoint $= 0.15$, steepness $= 5$), while the lowest ($0.106$) occurs at (midpoint $= 0.35$, steepness $= 50$). Higher collateral requirements impose friction on merchants and reduce overall participation of the market.

\textit{Loss reduction rate} ranges from $16\%$ to $44\%$, remaining positive throughout. The highest reduction ($44\%$) occurs at (midpoint $= 0.20$, steepness $= 10$), where collateral requirements are moderate enough to sustain adoption while still providing meaningful coverage. Higher steepness and higher midpoint both reduce the loss reduction rate, as aggressive or delayed collateral requirements suppress adoption and narrow the pool of covered transactions.

\textit{Failure reduction rate} is highest at low midpoint and low steepness (approximately $31\%$ at midpoint $= 0.10$, steepness $= 5$), where collateral requirements activate early and apply gradually across risk levels, deterring a broad range of risky transactions. Higher steepness concentrates deterrence around the midpoint, reducing its effectiveness for transactions outside that narrow risk band. Higher midpoint delays collateral activation, allowing more risky transactions to proceed unchecked and reducing failure deterrence to as low as $7\%$ at (midpoint $= 0.35$, steepness $= 50$).

\textit{Final wallet} is highest at low steepness and high midpoint: \emph{e.g.}, $11,166$ at (midpoint $= 0.35$, steepness $= 5$), and lowest at high steepness and low midpoint: \emph{e.g.}, $1,405$ at (midpoint $= 0.10$, steepness $= 50$). Lower midpoint activates collateral requirements at lower estimated risk, while higher steepness concentrates those requirements sharply around the midpoint; the combination of high midpoint and low steepness spreads collateral demands gradually across a wide risk range, improving solvency.

\paragraph{Takeaway.}
Across all sweeps, \textsc{ARS} realises a structured tradeoff: tighter underwriting through higher loading, lower FN, or more aggressive collateralisation generally improves underwriter solvency and often reduces user loss, but introduces friction that reduces adoption. Importantly, the failure reduction rate reveals a system-level benefit: the underwriter's collateral requirements deter risky transactions from executing, improving overall marketplace health beyond the direct protection of individual covered users. Critically, given an underwriter system, these tradeoffs are explicit and tunable through parameters such as $\lambda$, sigmoid midpoint and steepness, and no single parameter controls all four metrics independently. 

\section{Implications of \textsc{ARS} and Research Agenda}
\textsc{ARS} reframes end-to-end trust for agentic systems at the transaction layer. Rather than aiming to eliminate stochastic failures, it specifies how residual uncertainty can be handled through explicit fund-control semantics with escrow for contingent fees and risk underwriting for pre-verification principal exposure. In this framing, ``trust'' is operationalized as a verifiable settlement rule: when predefined outcome conditions fail, a predetermined economic remedy is triggered.

\paragraph{ARS is one assurance layer, not the only one.}
Economic assurance is not the only route to user-facing guarantees. \textit{Regulatory and legal mechanisms} such as liability allocation, disclosure, consumer protection, and enforcement can also provide deterministic remedies, albeit through institutional rather than standard-level execution. A key open question is how standard-native guarantees such as escrow and underwriting should interoperate with legal accountability: which failures should be handled by automated settlement, which require human adjudication, and which should be escalated to formal legal processes.

\paragraph{The bottleneck is risk modeling, not settlement mechanics.}
The central promise of \textsc{ARS} depends on whether an underwriter can estimate and price risk with sufficient accuracy to avoid undue friction. If the underwriter overestimates risk, high-quality providers face unnecessary collateral and increased friction when participating the market; if it underestimates risk, the underwriter becomes insolvent and the guarantee loses credibility. This shifts part of the trustworthy-agent problem from purely model-internal safety to a measurement problem: estimating failure probabilities and loss magnitudes under heterogeneous tasks, environments, and merchants.

Thus, \textsc{ARS} suggests several \textbf{concrete new research directions.}\\ 
\textbf{Loss modeling}: some failures map naturally to monetary loss, for example, unauthorized transfers, trading misexecution, non-delivery after prepayment, leakage of credentials with directly measurable impact. Other failures such as hallucination, biased outputs, defamation, privacy harm, or psychological impact also produce real harm but lack an agreed monetary metric. Studying how to define contractible, auditable proxies for these harms is necessary for extending assurance beyond purely well-defined losses in terms of financial loss.\\
\textbf{Frequency estimation and stress regimes}: underwriting requires not only identifying failure types but quantifying their incidence under realistic distributions of prompts, tool environments, and user behaviors. This motivates systematic measurement of failure rates as a function of scenario variables such as task complexity, tool access, time pressure, distribution shift, and the design of benchmarks that capture tail-risk behavior rather than average-case performance.\\
\textbf\emph{Mechanism design and incentives}: deposits and reimbursement rules change provider incentives and user adoption. Understanding how to set collateral, coverage limits, and premiums to balance adoption, solvency, and user protection is a central mechanism-design problem, especially under imperfect detectors and strategic behavior 

\paragraph{Relationship to model-centric trustworthy AI.}
\textsc{ARS} is not a substitute for foundational work on fairness, robustness, and alignment. Instead, it complements model-centric approaches by providing a transaction-layer interface for accountability. Improvements in model reliability reduce the frequency and severity of claimable failures, lowering required collateral and premiums, and increasing adoption. Conversely, \textsc{ARS} provides an operational lens for safety research by forcing explicit definitions of failure triggers, evidence requirements, and loss models, and by motivating quantitative evaluation of failure frequency and consequence under deployment-like conditions.

\section{Conclusion}
We introduced the \textbf{Agentic Risk Standard}, a settlement-layer service for end-to-end assurance in agentic services. ARS formalizes agentic tasks as jobs governed by a signed structured agreement and a deterministic lifecycle with a small typed action vocabulary. It separates \emph{service compensation} from \emph{execution principal}: service fees are handled via escrow-style conditional settlement, while fund-involving jobs optionally add an underwriting-backed principal track that gates pre-execution capital release through risk assessment, collateralization, adaptive multi-signature authorization, and explicit human override when users choose to proceed without underwriting.

This standard framing shifts trust from model-internal properties to enforceable transaction semantics, making user protection contractible and auditable across heterogeneous agents, marketplaces, and payment rails. In a simplified simulation study, we demonstrated that underwriting can materially reduce user loss while exposing clear and tunable tradeoffs among user protection, underwriter solvency, and merchant-side friction. These results motivate future work on risk modeling for diverse failure modes, empirical measurement of failure frequencies under deployment-like conditions, and the design of underwriting and collateral schedules that remain robust under detector error and strategic behavior.

\bibliography{colm2026_conference}
\bibliographystyle{colm2026_conference}

\end{document}